\documentclass[journal]{IEEEtran}
\usepackage{cite}
\usepackage[mathscr]{euscript}
\def\x{{\mathbf x}}

\usepackage{graphicx}
\usepackage[cmex10]{amsmath}
\usepackage[ruled]{algorithm2e}
\usepackage{algorithmic}
\usepackage{array}
\usepackage{xcolor}
\newcolumntype{P}[1]{>{\centering\arraybackslash}p{#1}}
\newcolumntype{M}[1]{>{\centering\arraybackslash}m{#1}}
\usepackage{mdwmath}
\usepackage{mdwtab}
\usepackage{url}
\usepackage[english]{babel}
\usepackage{amsfonts}
\usepackage{mathbbol}
\usepackage{amssymb}
\usepackage{amsthm}
\usepackage{mathtools}
\usepackage{adjustbox}
\usepackage{comment}
\usepackage{multicol}
\usepackage{multirow}
\usepackage{placeins}
\usepackage{dsfont}
\usepackage{soul,color}
\usepackage{tabu}
\usepackage{multirow}
\usepackage{booktabs}
\usepackage{cleveref}
\usepackage{cite}
\usepackage{paracol}
\usepackage[many]{tcolorbox}
\usepackage{xcolor}
\usepackage{physics}
\usepackage{circuitikz}
\usepackage{enumitem}
\usepackage{framed}
\usepackage{enumitem}
\usepackage{framed}
\usepackage{wrapfig}
\usepackage{url}

\NewTColorBox{NewBox}{ s O{!htbp}  }{%
  floatplacement={#2},
  IfBooleanTF={#1}{float*,width=\textwidth}{float},
  colframe=black!50!black,colback=white!95!black,valign=center, halign=flush center,enhanced,attach boxed title to top center={yshift=-3mm,yshifttext=-1mm},colbacktitle=black!70!white,
  title=Objectives,fonttitle=\bfseries,
  boxed title style={size=small,colframe=black!50!black}
  }
\usepackage{float}
\usepackage{etoolbox}
\makeatletter
\patchcmd{\@makecaption}
  {\\}
  {:\ }
  {}
  {}
\patchcmd{\@makecaption}
  {\scshape}
  {}
  {}
  {}
\makeatother

\newcommand*{\qeda}{\hfill\ensuremath{\blacksquare}}

\SetAlgoCaptionLayout{centerline}

\begin{document}

\title{Resonant Machine Learning Based on Complex Growth Transform Dynamical Systems }

\author{Oindrila~Chatterjee,~\IEEEmembership{Student~Member,~IEEE,}
        and~Shantanu~Chakrabartty,~\IEEEmembership{Senior~Member,~IEEE}

\thanks{Both the authors are with the Department of Electrical and Systems Engineering, Washington University in St. Louis, St. Louis, Missouri 63130, USA. All correspondences regarding this manuscript should be addressed to shantanu@wustl.edu.}
\thanks{This work was supported in part by a research grant from the National Science Foundation (ECCS:1550096).}}

\maketitle
\begin{abstract}
 
 Traditional energy-based learning models associate a single energy metric to each configuration of variables involved in the underlying optimization process. Such models associate the lowest energy state to the optimal configuration of variables under consideration, and are thus inherently dissipative. In this paper we propose an energy-efficient learning framework that exploits structural and functional similarities between a machine learning network and a general electrical network satisfying the Tellegen's theorem. In contrast to the standard energy-based models, the proposed formulation associates two energy components, namely, active and reactive energy to the network. This ensures that the network's active-power is dissipated only during the process of learning, whereas the reactive-power is maintained to be zero at all times. As a result, in steady-state, the learned parameters are stored and self-sustained by electrical resonance determined by the network's nodal inductances and capacitances. Based on this
approach, this paper introduces three novel concepts: (a) A learning framework where the network's active-power
dissipation is used as a regularization for a learning objective function that is subjected to zero total reactive-power
constraint; (b) A dynamical system based on complex-domain, continuous-time growth transforms which optimizes the learning objective function and drives the network towards electrical resonance under steady-state operation; and (c) An annealing procedure that controls the trade-off between active-power dissipation and the speed of convergence. As a representative example, we show how the proposed framework can be used for
designing resonant support vector machines (SVMs), where we show that the support-vectors correspond to an LC network with self-sustained oscillations. We also show that this resonant network dissipates less active-power compared to its non-resonant counterpart.

\end{abstract}


\section{Introduction}
\label{intro}

\IEEEPARstart{F}{rom} an energy point of view, the dynamics of a machine learning framework is analogous to an electrical network since both the networks evolve over a conservation manifold to attain a low-energy state. 
In literature, this analogy has served as the basis for energy-based learning models, where the learning objective function is mapped to an equivalent network energy.~\cite{lecun2006tutorial,hopfield1985neural}. The network variables then evolve according to some physical principles 
subject to network constraints to seek out an energy optimal state. 
\begin{figure}
\begin{center}
\includegraphics[page=1,scale=0.65]{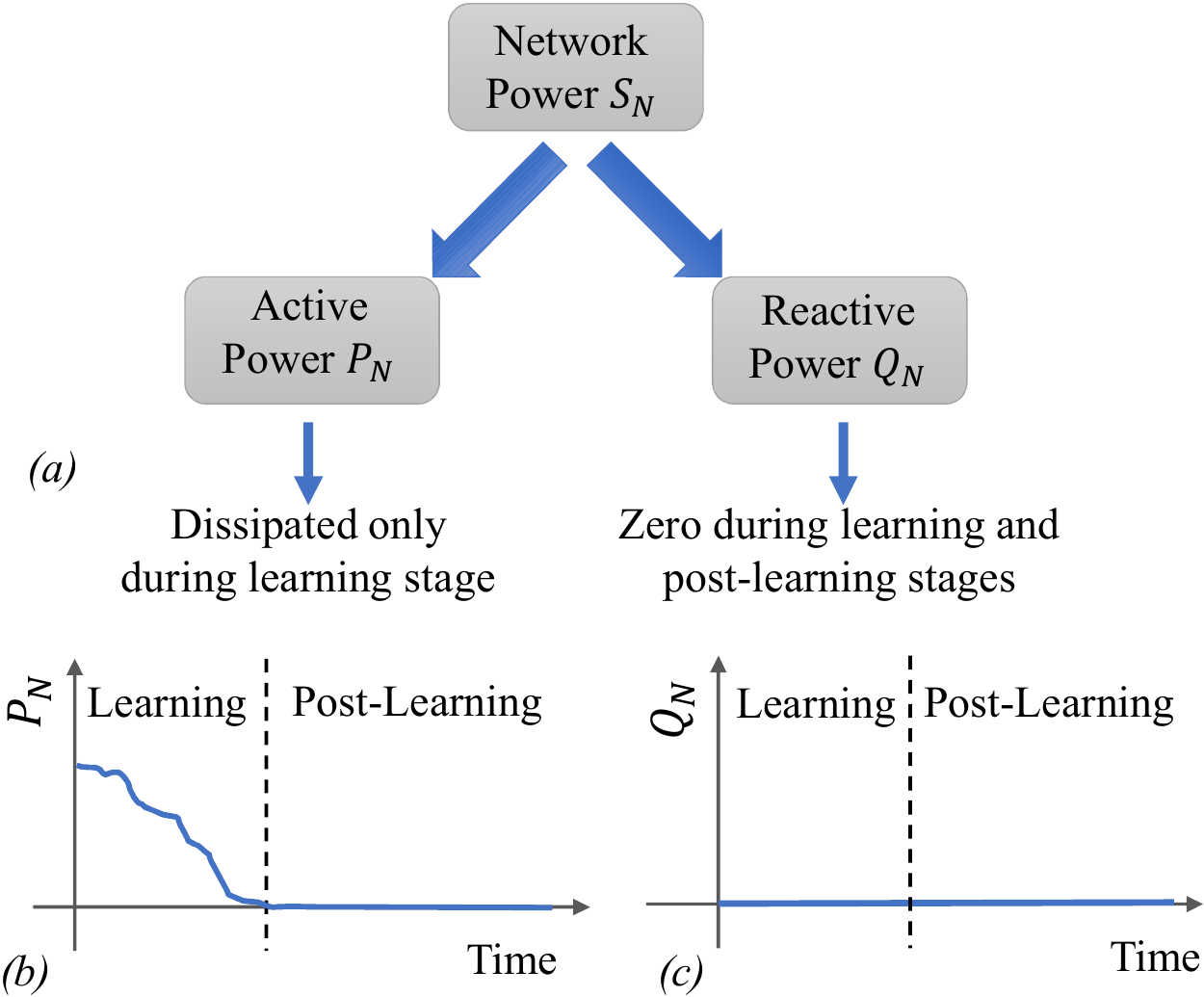}
\end{center}
\caption{(a) The total or apparent power $S_N$ of the electrical network comprising of the active-power $P_N$ or the dissipated power and the reactive-power $Q_N$ or the power used for energy-storage. The goal of the proposed resonant learning framework: (b) minimizing active-power $P_N$ during learning and ensuring $P_N=0$, post-learning or steady-state; and (c) maintaining $Q_N = 0$ in learning and post-learning phases.}
\label{fig_summary}
\end{figure}
However, even if the solution reached by such energy-based learning models may be optimal from the objective function point of view, it may not be the most energy-efficient solution when considering an actual physical implementation. This is because most of these formulations assume that the energy being minimized in the network is  dissipative in nature. Whereas, in a physical electrical network, the total power $S_N$ (also known as the apparent power) comprises not only of the dissipative component (also referred to as the active-power) but also a latent or stored non-dissipative component (also referred to as the reactive-power)~\cite{edminister1965theory,mac1964network}. 
This is illustrated in Figure \ref{fig_summary}(a) and can be mathematically expressed as:
\begin{align} \label{eq_networkenergy}
\text{Total Network Power} \enskip S_N=\text{Active Power} \enskip  P_N \nonumber \\ + j \times \enskip \text{Reactive Power} \enskip Q_N
\end{align}
where $j = \sqrt{-1}$ denotes the imaginary component.
While the active-power $P_N$ represents the rate-of-energy loss in the network, the reactive-power $Q_N$ represents the rate of change in stored energy in the network's electric and magnetic fields (typically modeled as lumped capacitive and inductive elements). In the design of electrical networks, reactive-power is generally considered to be a nuisance since it represents the latent power that does not perform any useful work~\cite{dixon2005reactive,hofmann2012reactive}. However, from the point of view of learning, the reactive-power could be useful not only for storing the learned parameters of the network, but could also improve the dynamics of the network during learning. In this paper, we propose such a framework which exploits both active and reactive network power for learning and memory. The objective will be to achieve a network power profile as
illustrated in Figures~\ref{fig_summary} (b) and (c). During learning, the network will optimize its active-power ($P_N$) and under steady-state condition or post-learning, ensure $P_N=0$. The reactive-power ($Q_N$),
on the other hand, will always be maintained at zero. This implies that the stored energy (mathematically - the time-integral of the reactive-power) is conserved across the learning and post-learning phases respectively. Thus, during the post-learning phase or in steady-state, the network will not dissipate any power, and the reactive energy is used to maintain its current network state or memory. 

This steady-state condition corresponds to a state of electrical resonance, and in Section \ref{sec_energymodel} we generalize this concept to a framework of resonant machine learning. To reach this steady-state, in Section \ref{sec_growth}, we present a dynamical system based on complex-domain continuous-time growth-transforms, which extends our previous work on growth transform networks using real-variables\cite{chatterjee2018decentralized}. The complex-domain formulation allows manipulation of the relative phase between the voltage and current variables associated with the network nodes and thus is used to optimize the active-power dissipation during learning. While the approach could be applied to different learning networks, in Section \ref{sec_learning}, we use this framework for designing resonant one-class support vector machines (SVMs) \cite{scholkopf2000support}. In this context, we also compare the performance of the resonant SVM model with its non-resonant variant on both synthetic and real datasets. Finally, Section \ref{sec_discussions} concludes the paper with a brief discussion on the implication of the proposed model when applied to other network-based models that do not involve learning, for instance, the coupled oscillator networks\cite{strogatz2000kuramoto}. 

 \par The key contributions of the paper can be summarized as follows:
\begin{itemize}
	\item Contrary to standard energy-based models which optimize for a single energy metric, we map a learning problem to an electrical network with two energy components: dissipative(active) and non-dissipative(reactive) energy.
	\item Active power dissipation is introduced as a regularizer in the original learning problem while enforcing zero total reactive power by driving the network into electrical resonance. This ensures zero active power dissipation in the post-learning stage. 
	\item We present a novel growth transform-based complex dynamical system for optimizing the cost function that ensures a resonant condition in steady-state.
	\item We propose an annealing schedule that can trade-off speed of convergence with active-power dissipation for different applications.   
\end{itemize}

\subsection{Notations}

\par For the rest of the paper, we will follow the mathematical notations summarized in Table \ref{notations} below:

\begin{table}[!htbp] 
	\renewcommand{\thetable}{\arabic{table}}
	\centering
	\caption{Notations}
	\label{notations}
	\renewcommand{\arraystretch}{2}
	\begin{tabular}{|c|l|} \hline
		\centering
		\textbf{Variable} & \textbf{Definition} \\
		\hline
		$\mathbb{R}_+$ & One-dimensional positive real vector space \\ 
		\hline
		$\mathbb{R}^N$ & N-dimensional real vector space \\ 
		\hline
		$\mathbb{C}$ & One-dimensional complex vector space \\  
		\hline
		
		$\lvert z \rvert$ & magnitude of a complex variable $z$\\
		\hline
		$\Re(z) $ & real part of a complex variable $z$\\
		\hline
		$\Im(z) $ & imaginary part of a complex variable $z$\\
		\hline
		$z^*$ & complex conjugate of a complex variable $z$\\ 
		\hline 
		$z(t)$ & a continuous-time complex variable at time $t$ \\
		\hline
		$z_{n}$ & a discrete-time complex variable at the $n^{\text{th}}$ time step \\
		\hline
	\end{tabular}
\end{table}

\section{Background and Related works}
\label{sec_background}
\subsection{Energy-based Learning Models}
\label{subsec_energybased}

The principle of minimum energy states that all self-organizing physical systems evolve dynamically over an intrinsic conservation manifold to attain their minimum energy configuration in the steady state\cite{feynman2017character}. Energy-based machine learning models follow a similar cue, where the goal is to find the optimal configuration of a predetermined energy landscape determined by the learning problem at hand\cite{lecun2006tutorial,boureau2007unified}. Some of the most notable examples of energy-based models are those based on the Ising model of statistical mechanics, like the Hopfield network\cite{hopfield1985neural}, and its stochastic counterpart the Boltzmann machine \cite{hinton1983optimal} and its variants\cite{salakhutdinov2007restricted}. In these models, the network converges to the local minimum of a Lyapunov function in steady state. Another class of energy-based models (EBMs)\cite{lecun2006tutorial} propose a unified learning framework by mapping the learning problem to a global scalar ``energy function". The algorithm minimizes a loss functional to seek out an optimal energy landscape that associates lowest energy to the observed configuration of variables involved in the underlying optimization process. These methods are essentially non-probabilistic, and often involve intractable integrals which require MCMC methods for normalization when applied to probabilistic frameworks. Different classes of optimization algorithms like contrastive divergence, contrastive hebbian learning, equilibrium propagation etc.\cite{scellier2017equilibrium} have been proposed for a variety of supervised, unsupervised and reinforcement learning applications\cite{ngiam2011learning,zhai2016deep,haarnoja2017reinforcement,lawson2019energy} in the context of energy-based models. However, all of these approaches consider an energy metric that solely depends on the learning task under consideration. This implies that in an analog implementation of the model, the optimal solution may not be the most energy-efficient one.

\begin{figure*}
	\begin{center}
		\includegraphics[page=1,scale=0.65]{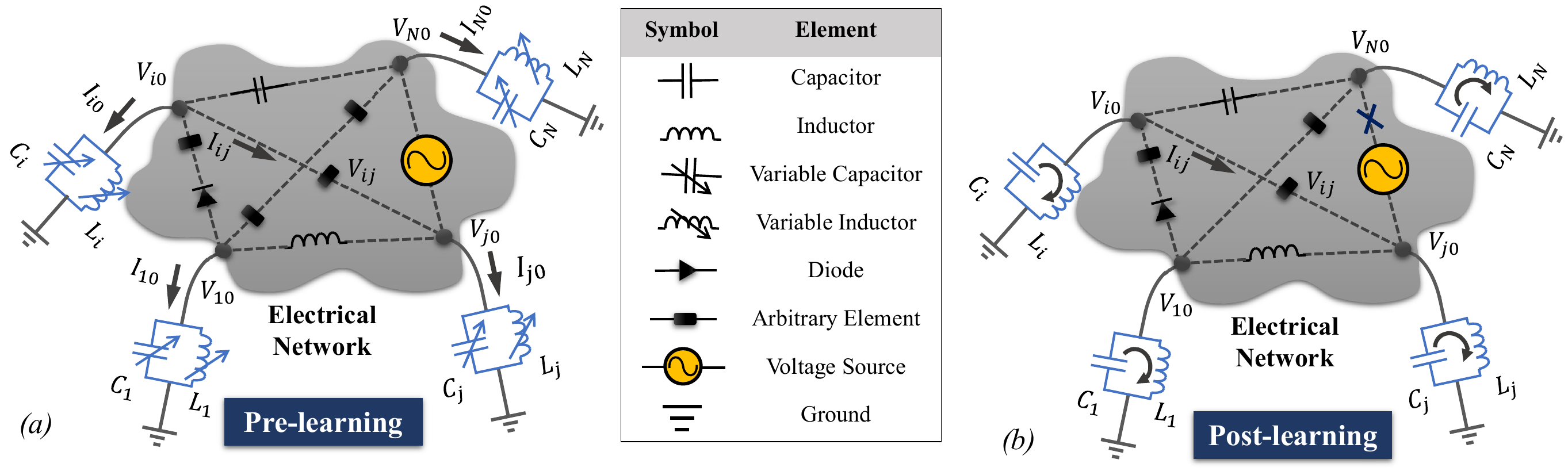}
	\end{center}
	\caption{Equivalent network model comprising of $N$ electrical nodes, with an inductance and capacitance element
		associated with each of the nodes. (a) Learning is equivalent to changing the values of inductive and capacitive elements; and (b) in steady-state the network is driven into electrical resonance.}
	\label{fig_motivation}
\end{figure*}

\subsection{Complex-domain Machine Learning Models}
\label{subsec_complexlearning}
A variety of complex domain learning models have been proposed in literature for different learning algorithms, e.g., complex neural networks, complex support vector machines, complex deep networks etc\cite{hirose2003complex,trabelsi2017deep,gaudet2018deep,guberman2016complex,bouboulis2014complex}. In addition to providing the phase variables which allow for another degree of freedom, other advantages of the complex domain operation have been shown in these works. It has been demonstrated that complex learning models lead to a richer set of dynamics, noise robustness  and better convergence properties in the case of classification problems\cite{trabelsi2017deep,bruna2015mathematical,nitta2004orthogonality}. Moreover, phase information might provide additional insights in the context of many complex-valued physical signals (or complex domain transforms of physical signals, e.g., Fourier or wavelet transforms)\cite{trabelsi2017deep,nakashika2017complex}. However, most of these complex learning models either treat the real and imaginary components of the signal separately, or do not utilize the entire phase space when operating in the complex domain.

\section{Optimization and Electrical Resonance}\label{sec_energymodel}

\begin{figure}
	\begin{center}
		\includegraphics[page=1,scale=0.7]{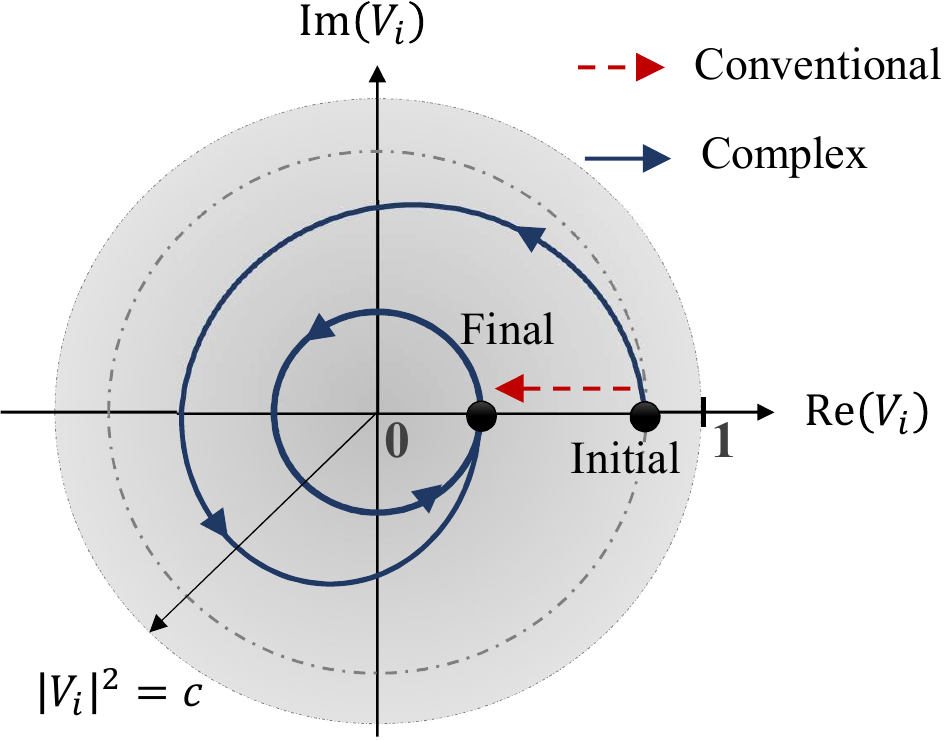}
	\end{center}
	\caption{Illustration showing that operating in the complex domain allows different possible learning trajectories from an initial state to the final steady-state. Regularization with respect to the phase-factor could then be used to select the trajectory with an optimal active-power dissipation profile, and results in limit cycle oscillations in steady-state. The circles indicate the constant magnitude loci.}
	\label{fig_learningrep}
\end{figure}


\par Consider an electrical network as shown in Figure~\ref{fig_motivation}(a), comprising of $N$ internal nodes. 
The voltage difference between the $i^{\text{th}}$ and $j^{\text{th}}$ nodes is denoted by $V_{ij}$, with $V_{i0}$ being the $i^{\text{th}}$ nodal voltage with respect to
ground terminal (referred to as the $0^{\text{th}}$ node). Similarly, the current flowing between the $i^{\text{th}}$ and $j^{\text{th}}$ nodes is given by $I_{ij}$, and $I_{i0}$ denotes the current flowing out of the $i^{\text{th}}$ node into the ground terminal. Then, according to the Tellegen's theorem~\cite{tellegen1952general}:
\begin{equation}
\sum \limits_{i} V_{ij}I_{ij}^*=0 ,
\end{equation}
which states that the total complex electrical power or apparent power is zero. Isolating the apparent power flowing from the nodes to the ground terminal from that flowing between the internal nodes, we have:
\begin{gather}
\sum \limits_{i \neq j,0;j \neq 0}(V_{ij})I_{ij}^*+\sum \limits_{i}(V_{i0})I_{i0}^*=0 \nonumber \\
\implies \sum \limits_{i=1}^N V_{i0}I_{i0}^*=-\sum_{i,j=1}^N V_{ij}I_{ij}^* \nonumber \\
\implies S_T = -S_N \nonumber \\
\implies P_T+jQ_T=-P_N-jQ_N \nonumber \\
\implies \lvert P_T \rvert  = \lvert P_N \rvert \nonumber \\
\implies \lvert Q_T \rvert  = \lvert Q_N \rvert
\label{eq_tellegen}
\end{gather}
where $S_T=\sum_i  V_{i0}  I_{i0}^*$ is the nodal apparent power, and $P_T=\sum_i \Re\{ V_{i0}  I_{i0}^*\}$ and $Q_T=\sum_i \Im\{ V_{i0}  I_{i0}^*\}$ are the total active and reactive power consumed at the nodes.  Similarly, $S_N$, $P_N$ and $Q_N$ represent the apparent, active and reactive power consumed due to current flow between the network nodes (other than the ground terminal). Note that this result holds even if active-power sources are embedded in the network, as shown in Figure~\ref{fig_motivation}(a). Thus, Equations~(\ref{eq_tellegen}) imply that if we minimize the active-power at the nodes of the network $P_T$ subject to the constraint that the nodal reactive power $Q_T=0$, then the formulation would be equivalent to minimizing the network active-power $P_N$ while ensuring that the network reactive power $Q_N=0$.
 This result can be expressed as:
\begin{align}
\underset{\{ V_i, I_i  \in \mathbb{C}\}} {\text{min}} \quad  \mathcal{D}=\sum \limits_{i=1}^N \lvert \Re\{V_iI_i^*\} \rvert^2  \nonumber  \\
\textit{s.t.}\quad \sum_i \Im\{ V_{i}  I_{i}^*\}=0 \label{eq_conservationconst_comp1}
\end{align}
where we have used the notations $V_{i0}=V_i$ and $I_{i0}=I_i$ for the sake of brevity.
If we assume that  the  $i^{\text{th}}$ node is associated with a lumped capacitance $C_i$, and a lumped inductance $L_i$,  ensuring zero nodal reactive-power implies
\begin{equation}
\sum \limits_{i=1}^N V_i \Big(C_i \dfrac{dV_i}{dt}\Big)^*+\Big(L_i\dfrac{dI_i}{dt}\Big)I_i^*=0, \label{eq_resonant}
\end{equation}
where $\Big(C_i \dfrac{dV_i}{dt}\Big)$ and $\Big(L_i\dfrac{dI_i}{dt}\Big)$ represent the current flowing through $C_i$ and the voltage across $L_i$ respectively. 
Equation (\ref{eq_resonant}) is equivalent to: 
 \begin{align}
\sum_{i=1}^N \Big(\dfrac{1}{2}C_i \lvert V_i\rvert ^2+\dfrac{1}{2}L_i\lvert I_i\rvert ^2\Big)=E_0, \label{eq_conservationconst}
\end{align}
where $\lvert V_i\rvert ^2=V_iV_i^*$ and $\lvert I_i\rvert ^2=I_iI_i^*$, implying that the total network reactive energy is conserved to be equal to some constant value $E_0$. Satisfying
this constraint is equivalent to sustaining a condition of electrical resonance. The optimization problem (\ref{eq_conservationconst_comp1}) can be transformed as: 
\begin{align}
\underset{\{\lvert V_i \rvert,\lvert I_i \rvert \in \mathbb{R}_+,\phi_i \in \mathbb{R} \}} {\text{min}} \quad  \mathcal{D}=\sum \limits_{i=1}^N \lvert V_i \rvert^2 \lvert I_i \rvert^2 \cos^2\phi_i    \label{eq_conservation}\\
\textit{s.t.}\quad \sum_{i=1}^N \Big(\dfrac{1}{2}C_i \lvert V_i\rvert ^2+\dfrac{1}{2}L_i\lvert I_i\rvert ^2\Big)=E_0, \enskip \lvert \phi_i \rvert \le \pi \enskip \forall i \label{eq_conservationconst}
\end{align}
 where $\phi_i$ denotes the phase-angle between the voltage and current phasors at the $i^{\text{th}}$ node. Note that the optimization in Equation~(\ref{eq_conservation}) admits
only three types of solutions in steady-state: (a) ($\lvert I_i\rvert \neq 0, \lvert V_i\rvert \neq 0, \lvert \phi_i \rvert = \pi/2$) which corresponds to a resonant LC tank; (b) ($\lvert I_i\rvert = 0, \lvert V_i\rvert \neq 0$)  which corresponds to an electrically isolated or floating node; and (c) ($\lvert I_i\rvert \neq 0, \lvert V_i\rvert = 0$)  which corresponds to a short-circuit. In Appendix A we illustrate steady-state resonance conditions using a simple LC tank. Note that in all cases, active power is dissipated only during the learning phase, where $C_i$ and $L_i$ adapt to change the relative
magnitude of the voltage and current variables. Figure \ref{fig_motivation}(b) illustrates this resonant condition of the network in the post-learning stage, whereby active power sources in the network get electrically isolated from the rest of the network. The lumped capacitance and inductance associated with the LC network at the terminal nodes adapt such that the resonant frequency condition in maintained in steady-state (see Appendix A). 
This implies that in steady-state, the learned network parameters are stored and sustained by the resonant electric and magnetic fields of the LC tanks.
\par The constraint in Equation~(\ref{eq_conservationconst}) can be simplified by normalizing with respect to $E_0$ such that:
\begin{equation} \label{eq_normconservation}
\sum\limits_{i=1}^N \Big(\lvert V_i \rvert ^2+\lvert I_i \rvert ^2\Big)=1,
\end{equation}
where $V_i \leftarrow \sqrt{\dfrac{C_i}{2E_0}}V_i$ and $I_i \leftarrow \sqrt{\dfrac{L_i}{2E_0}}I_i$ represent the dimension-less voltages and currents. Henceforth, unless stated otherwise, we will use dimension-less quantities in our derivations.  

\begin{figure*}
\begin{center}
\includegraphics[page=1,scale=0.65]{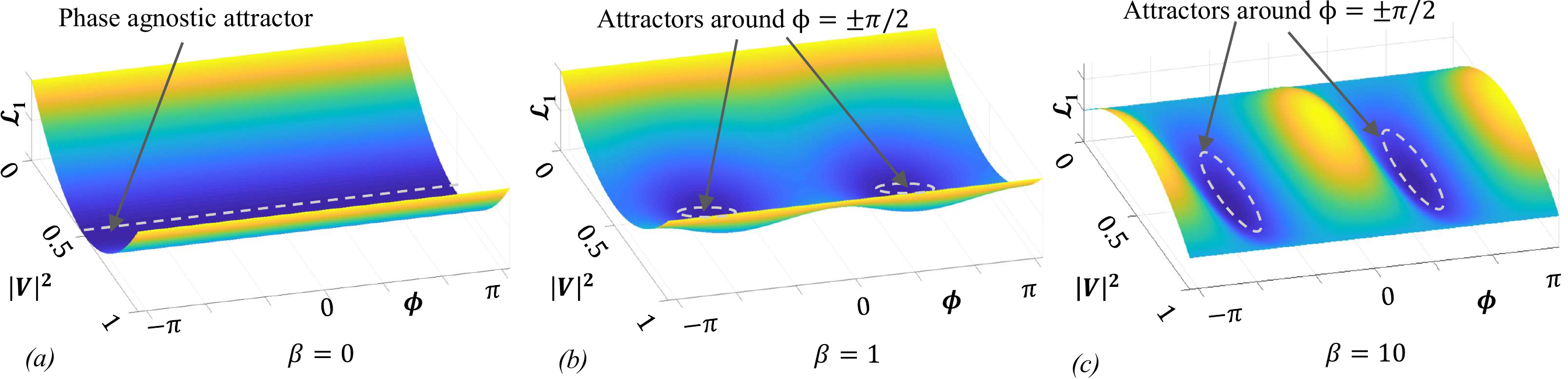}
\end{center}
\caption{Cost function $\mathcal{L}_1$ plotted for different values of the hyperparameter $\beta$: (a) $\beta=0$; (b) $\beta=1$ and (c) $\beta=10$.  }
\label{fig_modcost}
\end{figure*}
 
\par We now extend the optimization framework in Equation~(\ref{eq_conservation}) to include a general optimization function $\mathcal{H}$ as:
\begin{align}
\label{eq_optimclass}
\underset{\{\lvert V_i \rvert,\lvert I_i \rvert,\phi_i\}} {\text{min}} \quad  \mathcal{L}(\{\lvert V_i \rvert,\lvert I_i \rvert,\phi_i\})=\mathcal{H}(\{\lvert V_i \rvert,\lvert I_i \rvert\})
+\beta \mathcal{D} \nonumber \\
\textit{s.t.}\quad \sum_{i=1}^N \Big( \lvert V_i \rvert ^2+\lvert I_i \rvert ^2 \Big ) = 1, \enskip \lvert \phi_i \rvert \le \pi \enskip \forall i,
\end{align}
In this formulation, the active-power dissipation $\mathcal{D}$ in Equation~(\ref{eq_optimclass}) acts as a regularization function with $\beta \ge 0$ being a hyper-parameter. 
Note that the objective function $\mathcal{H}(\{\lvert V_i \rvert,\lvert I_i \rvert\})$ is only a function of the magnitudes of the
voltages and currents and is independent of the phase-angle $\phi_i$. This ensures independent control of the magnitudes and the phase to achieve the desired objective of optimizing the active-power dissipation. This is illustrated in Figure \ref{fig_learningrep} where controlling the phase allows different paths from the initial to the final state,
whereas evolution over the real-domain allows only one possible path. The complex-domain approach thus results in steady-state limit cycle oscillations that encode the final solution. Compared to other complex-domain machine learning frameworks~\cite{trabelsi2017deep,bouboulis2014complex}, the proposed formulation avoids
non-linear processing of phase/frequency which produces unwanted higher-order harmonics. This would have made it difficult
to maintain the network in the state of resonance (at specific set of frequencies) under steady-state. 
 
\par The two important properties of the optimization framework in Equation (\ref{eq_optimclass}) are as follows:
\newline $\bullet$ For a convex cost function $H$, we have
\begin{equation} \label{eq_convexmin}
\underset{\{\lvert V_i \rvert,\lvert I_i \rvert,\phi_i\}} {\text{min}} \quad  \mathcal{L}(\{\lvert V_i \rvert,\lvert I_i \rvert,\phi_i\})= \underset{\{\lvert V_i \rvert,\lvert I_i \rvert\}} {\text{min}} \quad \mathcal{H}(\{\lvert V_i \rvert,\lvert I_i \rvert\})
\end{equation}
This result follows from the three possible solutions of optimization problem (\ref{eq_conservation}).
\newline $\bullet$ If $\beta$ is slowly annealed to a sufficiently high value, then $\phi_i \rightarrow \pi/2$ under steady state for $i:\lvert V_i \rvert \lvert I_i \rvert \neq 0$. This implies that network achieves zero active power dissipation in steady state.  
Note that the method also holds for non-convex objective functions as well. In this case however, the network might show resonant behavior at a locally optimal solution. 


 \textbf{Example 1:} Consider a single-variable quadratic optimization problem of the form $\mathcal{H}_1(x)= x^2$, subject to the constraint $ \lvert x \rvert \le 1, \enskip x \in \mathbb{R} $. Substituting $x=\lvert V \rvert ^2-\lvert I \rvert ^2$, the problem can be mapped (please see Appendix B for more details) into the form equivalent to Equation (\ref{eq_optimclass}) as: 
\begin{gather}
\underset{\{\lvert V \rvert, \lvert I \rvert,\phi\}} {\text{min}} \quad \mathcal{L}_1=(\lvert V \rvert ^2-\lvert I \rvert ^2)^2 + \beta \lvert V\rvert^2\lvert I\rvert^2\cos^2\phi \label{eq_example1} \\
\textit{s.t.}\quad  \lvert V \rvert ^2+\lvert I \rvert ^2  = 1,\quad  \lvert \phi\rvert \le \pi  
\label{eq_constraint1}
\end{gather}

Figures \ref{fig_modcost}(a)-(c) plots $\mathcal{L}_1$ for different values of $\beta$. As shown in~\ref{fig_modcost}(a),
and as expected for $\beta=0$, the cost function has several minima (or attractors), whereas for $\beta > 0$, the minima corresponds to $\phi= \pm \pi/2$, for which the active-power dissipation is zero. The Figures~\ref{fig_modcost}(b)-(c) show that controlling $\beta$ will control the optimization landscape (without changing the location of the attractors) and will determine the attractor trajectory. This feature has been exploited in later sections to optimize the active-power
dissipation profile during the learning phase. 
\setcounter{algocf}{1}
\begin{algorithm*}[!htbp]
\caption{Complex growth transform dynamical system (Proof in Appendix C)} 
\label{algo1}
\begin{algorithmic}
\vspace{0.2cm}
\STATE $\bullet$ For an optimization problem of the form: 
\begin{align}
\label{eq_main1}
\underset{\{\lvert V_i \rvert,\lvert I_i \rvert,\phi_i\}} {\text{min}} \quad  \mathcal{L}(\{\lvert V_i \rvert,\lvert I_i \rvert,\phi_i\})&=\mathcal{H}(\{\lvert V_i \rvert,\lvert I_i \rvert\})
+\beta \sum \limits_{i=1}^N \lvert V_i \rvert^2 \lvert I_i \rvert^2 \cos^2\phi_i \nonumber \\
\textit{s.t.}\quad \sum_{i=1}^N \Big( \lvert V_i \rvert ^2+\lvert I_i \rvert ^2 \Big ) &= 1, \enskip \lvert \phi_i \rvert \le \pi \enskip \forall i=1,\ldots,N, \enskip \beta \ge 0
\end{align}
\\
\vspace{0.1cm}
\STATE $\bullet$ If $\mathcal{H}(\{\lvert V_i \rvert,\lvert I_i \rvert\})$ is Lipschitz continuous in the domain $D=\{\lvert V_i \rvert,\lvert I_i \rvert:\sum_{i=1}^N \Big( \lvert V_i \rvert ^2+\lvert I_i \rvert ^2 \Big ) = 1 \}$, the following system of nonlinear dynamical equations
\begin{align}
\dfrac{\partial V_i(t)}{\partial t}& =j\omega\sigma_{V_i}(t)V_{i}(t)-\Delta \sigma_{V_i}(t)V_i(t), \label{eq_mainvol} \\
\dfrac{\partial I_i(t)}{\partial t}& =j(\omega+\omega_{\phi_i})\sigma_{I_i}(t)I_{i}(t)-\Delta \sigma_{I_i}(t)I_i(t), \label{eq_maincurr} \\
\text{and} \enskip \tau_i \omega_{\phi_i} +\phi_{i}(t) &=g_{\phi_i}(t) , \enskip \forall i=1,\ldots,N\\
\text{ensures that} \enskip \dfrac{\partial \mathcal{L}}{\partial t} \leq 0,
 \label{eq_mainphase}
\end{align}
\vspace{0.1cm}
where $\sigma_{V_i}(t)=\sqrt{\dfrac{1}{V_{i}^*\eta}\Big(-\dfrac{\partial \mathcal{L}}{\partial  V_i}+\lambda V_i^*\Big)}$, $\sigma_{I_i}(t)=\sqrt{\dfrac{1}{I_{i}^*\eta}\Big(-\dfrac{\partial \mathcal{L}}{\partial  I_i}+\lambda I_i^*\Big)}$, $\omega_{\phi_i}=\dfrac{d \phi_i(t)}{d t}$, $\Delta \sigma_{V_i}(t)=1-\sigma_{V_i}(t)$, $\Delta \sigma_{I_i}(t)=1-\sigma_{I_i}(t)$  and $g_{\phi_i}(t)=\pi \dfrac{\lambda \phi_{i}-\pi \dfrac{\partial \mathcal{L}}{\partial \phi_{i}}}{-\phi_{i}\dfrac{\partial \mathcal{L}}{\partial \phi_{i}}+\lambda \pi}$, with $\eta= \sum\limits_{k=1}^N \Bigg(V_{k}\Big[-\dfrac{\partial \mathcal{L}}{\partial V_{k}}+\lambda V_{k}^*\Big]
+I_{k}\Big[-\dfrac{\partial \mathcal{L}}{\partial I_{k}}+\lambda I_{k}^*\Big]\Bigg)$, $\omega$ is an arbitrary angular frequency, and $\tau_i$ is the time-constant associated with the evolution of $\phi_i$.
\\
\vspace{0.2cm}
\end{algorithmic}
\end{algorithm*}

\section{Complex Growth Transforms}\label{sec_growth}
The problem in Equation (\ref{eq_optimclass}) involves complex phasors and hence entails the use of learning models operating in the complex domain for reaching the optimal solution. To this end, in this section, we propose a dynamical system that can be used to solve this optimization problem. The main result is summarized in Table~\ref{algo1} and the details of the proof is given in Appendix C. 
 
\begin{figure*}
\begin{center}
\includegraphics[page=1,scale=0.65]{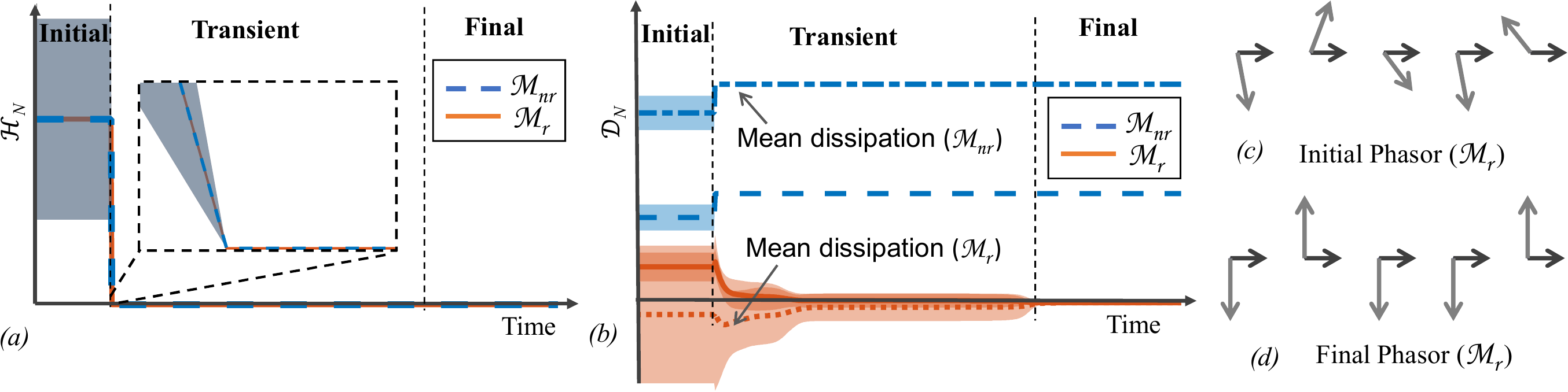}
\end{center}
\caption{Comparison of the resonant optimization model ($\mathcal{M}_{r}$) with its  non-resonant variant ($\mathcal{M}_{nr}$) for a quadratic objective function $\mathcal{L}_N$ shown in Equation~(\ref{eq_example2}). For this experiment, $N=5$, input frequency $\omega=\pi/10$, and the mean response is estimated over 10 trials with random initializations of $V_i,I_i,\phi_i \enskip \forall i=1,\ldots,5$: (a) comparison of the time-evolution of $\mathcal{H}_N$ for $\mathcal{M}_{nr}$ and $\mathcal{M}_{r}$, (b) comparison of the time-evolution of $\mathcal{D}_N$ ($\sum \limits_{i=1}^N \lvert V_i \rvert^2 \lvert I_i \rvert^2$ for $\mathcal{M}_{nr}$ and $ \sum \limits_{i=1}^N \lvert V_i \rvert^2 \lvert I_i \rvert^2 \cos^2 \phi_i$ for $\mathcal{M}_r$). For all the curves, the line indicates the mean response, while the shaded region shows the standard deviation about the mean across the 10 trials. The true dissipation ($\sum \limits_{n=1}^N \lvert V_i \rvert  \lvert I_i \rvert$ for $\mathcal{M}_{nr}$ and $\sum \limits_{n=1}^N \lvert V_i \rvert  \lvert I_i \rvert \cos \phi_i$ for  $\mathcal{M}_{r}$) over the 10 trials are also shown in the figure;   Phasor representations of the LC tank voltages and currents for a single trial : (c) initial configuration and (d) final configuration for $\mathcal{M}_{r}$. For $\mathcal{M}_{r}$, $\beta=1$ for all the trials.}

\label{fig_exampleresults}
\end{figure*}

\textbf{Theorem 1:} \textit{The system of nonlinear dynamical equations given by Equations (\ref{eq_mainvol})-(\ref{eq_mainphase})  in Table \ref{algo1} converge to the optimal point of Equation (\ref{eq_main1}) in the steady state, with zero energy dissipation i.e., $\sum \limits_{n=1}^N \lvert V_i \rvert  \lvert I_i \rvert \cos \phi_i=0$ (Proof given in Appendix C) .}
Consider an optimization framework which is an multi-variable extension of Example 1, given by Equations ~(\ref{eq_example1})-(\ref{eq_constraint1}):
\begin{gather}
\underset{\{\lvert V_i \rvert, \lvert I_i \rvert,\phi_i\}} {\text{min}} \quad \mathcal{L}_N=\sum \limits_{i=1}^N(\lvert V_i \rvert ^2-\lvert I_i \rvert ^2)^2 + \beta \sum \limits_{i=1}^N\lvert V_i\rvert^2\lvert I_i \rvert^2\cos^2\phi_i \label{eq_example2} \\
\textit{s.t.}\quad \sum \limits_{i=1}^N \Big( \lvert V_i \rvert ^2+\lvert I_i \rvert ^2 \Big)  = 1,\quad  \lvert \phi_i\rvert \le \pi \enskip \forall i  
\label{eq_constraint2}
\end{gather}
The optimal solution is reached when $\phi_i =\pm \frac{\pi}{2} \enskip \forall i$, which implies $\mathcal{L}_N=0$.  For the sake of comparison, we will consider two variants:  (a) the non-resonant model ($\mathcal{M}_{nr}$) where $\beta=0$, and (b) the resonant model ($\mathcal{M}_{r}$), where $\beta \neq 0$. Figures \ref{fig_exampleresults}(a) and (b) show a comparison of the cost $\mathcal{H}_N$ and the regularization metric $\mathcal{D}_N$ ($\sum \limits_{i=1}^N \lvert V_i \rvert^2 \lvert I_i \rvert^2$ for $\mathcal{M}_{nr}$ and $\sum \limits_{i=1}^N \lvert V_i \rvert^2 \lvert I_i \rvert^2 \cos^2 \phi_i$ for $\mathcal{M}_r$), for $N=5$ and $\omega=\pi/10$. In the case of $\mathcal{M}_{nr}$, $\mathcal{L}_N=\mathcal{H}_N$ and in the case of $\mathcal{M}_{r}$,  $\mathcal{L}_N=\mathcal{H}_N+\beta \mathcal{D}_N$, with $\beta=1$. The solid line indicates the mean response across $10$ trials with random initializations of $V_i,I_i$ and $\phi_i$. The shaded region shows the standard deviation across the trials in each case. Also shown in Figure \ref{fig_exampleresults}(b) are the true active-power dissipation profiles ($\sum \limits_{n=1}^N \lvert V_i \rvert  \lvert I_i \rvert$ for $\mathcal{M}_{nr}$ and $\sum \limits_{n=1}^N \lvert V_i \rvert  \lvert I_i \rvert \cos \phi_i$ for  $\mathcal{M}_{r}$) over the 10 trials.
\par It can be observed that under steady-state conditions, the model $\mathcal{M}_{nr}$ dissipates power. However for the model $\mathcal{M}_{r}$ the steady-state active-power goes to zero. This is illustrated in Figure \ref{fig_exampleresults}(b). Figure \ref{fig_exampleresults}(c) shows the initial phasor configuration for the currents and voltages at each node of the network for a single trial for $\mathcal{M}_r$. Here, we assume the voltage phasor $V_i$ to be aligned with the horizontal axis, and the current phasor $I_i$ to be at an angle $\phi_i$ with respect to the horizontal, $\forall i$. Figure \ref{fig_exampleresults}(d) shows the steady-state phasor configuration for $\mathcal{M}_{r}$ for the same trial. The voltage and current phasors are orthogonal to each other for all the nodes, thus verifying the zero active-power dissipation.

\par
\begin{figure}
\begin{center}
\includegraphics[page=1,scale=0.28,trim=4 4 4 4,clip]{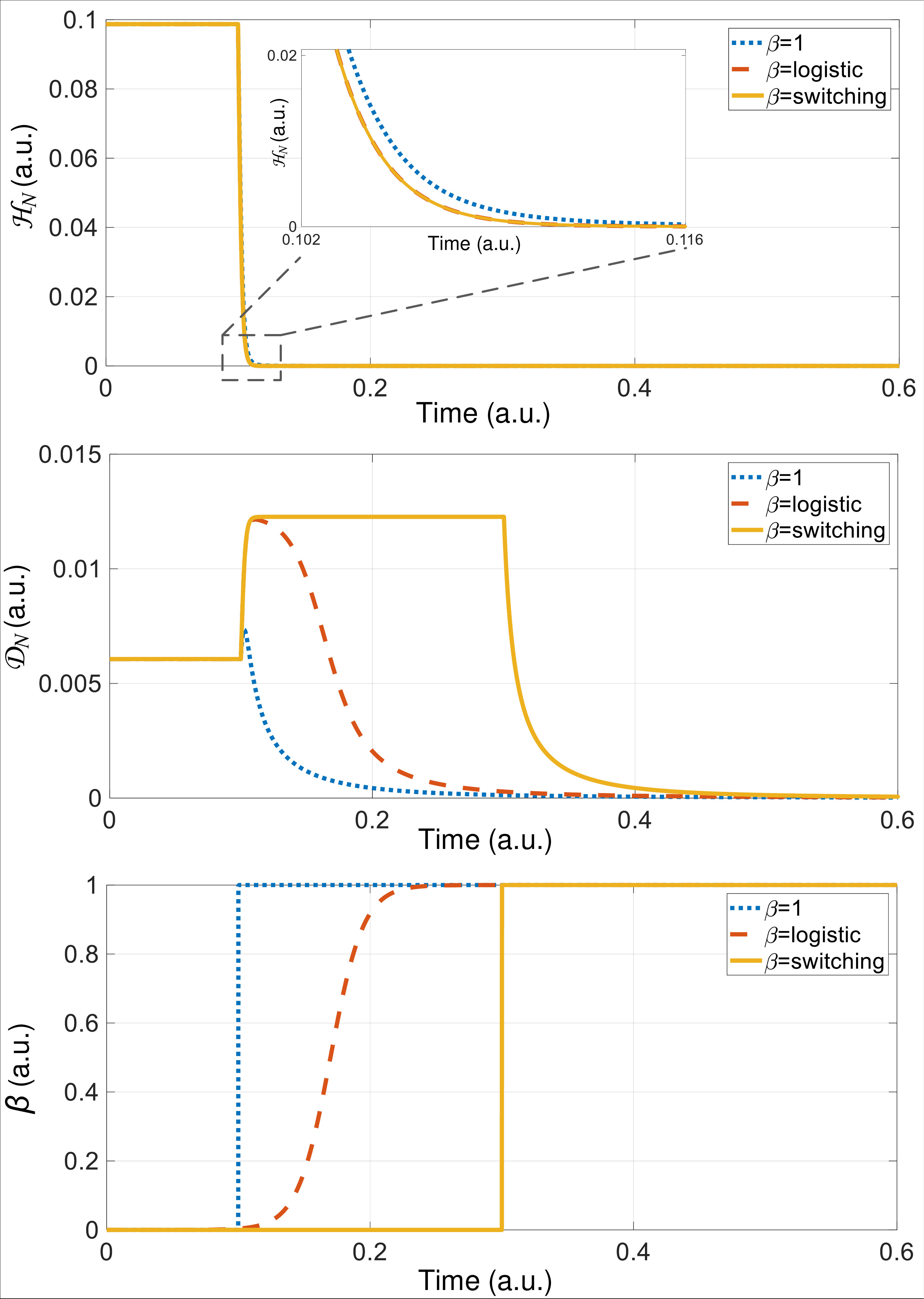}
\end{center}
\caption{Comparison of the performance of the resonant model $\mathcal{M}_{r}$ for different choices of annealing schedule for $\beta$ ($N=5,\omega=\pi/10$): (a) time evolution of $\mathcal{H}_N$ (inset shows a zoomed in view of the cost evolution in the transient phase), (b) time evolution of $\mathcal{D}_N$ and (c) time evolution of $\beta$. In all the cases, the optimization process starts after $0.1 \enskip a.u.$ from the onset of the simulation. The curves corresponding to $\beta=1$ denotes the case when $\beta$ takes a constant value from $t=0.1 \enskip a.u.$; $\beta=\text{logistic}$ corresponds to the cases when $\beta$ is slowly increased following a logistic curve from $t=0.1 \enskip a.u.$, and takes on a maximum value of $\beta=1$; $\beta=\text{switching}$ corresponds to the case when  $\beta$ switches from a minimum value($=0$) to a maximum value($=1$) at $t=0.3 \enskip a.u.$, after the system has converged to the optimal solution.}
\label{fig_examplecompare}
\end{figure}

 In the next set of experiments, we annealed the hyperparameter $\beta$ and evaluated its impact on the active-power dissipation metric $\mathcal{D}_N$ and the convergence of the object function $\mathcal{H}_N$ for the model $\mathcal{M}_{r}$. Figure \ref{fig_examplecompare} presents a comparison of the performance of $\mathcal{M}_{r}$ for different choices of annealing schedule for $\beta$, with the angular frequency $\omega=\pi/10$ as before. Figure \ref{fig_examplecompare}(a) shows the time evolution of the objective function $\mathcal{H}_N$, \ref{fig_examplecompare}(b) shows the time evolution of the dissipation metric $\mathcal{D}_N$ and \ref{fig_examplecompare}(c) shows the annealing schedules adopted for $\beta$. In all the cases, the optimization process starts after time $t=0.1 \enskip a.u.$ from the onset of the simulation. The curves corresponding to $\beta=1$ denote the case when $\beta$ takes a constant value from $t=0.1 \enskip a.u.$; $\beta=\text{logistic}$ corresponds to the case when $\beta$ is slowly increased from $\beta_{\text{min}}=0$ following a logistic curve of the form $\beta(t)=\beta_{\text{min}}+\dfrac{\beta_{\text{max}}-\beta_{\text{min}}}{1+\exp(-k(t+t_0))}$ from $t=0.1 \enskip a.u.$, and takes on a maximum value of $\beta_{\text{max}}=1$ ($k$ and $t_0$ are hyperparameters determining the steepness and mid-point of the sigmoid respectively); $\beta=\text{switching}$ corresponds to the case when  $\beta$ switches from a minimum value ($\beta_{\text{min}}=0$) to a maximum value ($\beta_{\text{max}}=1$) at $t=0.3 \enskip a.u.$, after the system has converged to the optimal solution. We can observe that in all of the cases, the model converges to the optimal solution, irrespective of the choice of $\beta$. However, different annealing schedules for $\beta$ lead to different active-power dissipation profiles. For example, a constant value of $\beta$ throughout the duration of the experiment would lead to faster minimization of the active-power dissipation metric, but at the cost of slower convergence. The opposite trend can be seen when $\beta$ is slowly annealed to a sufficiently high value throughout the course of the optimization. The annealing schedule thus acts as a trade-off between the speed of convergence, and the rate of minimization of the active power.

\subsection{Model Properties and Extensions}
The dynamical system represented by Equations (\ref{eq_mainvol})-(\ref{eq_mainphase}) and the resonant optimization framework also exhibits the following properties and extensions:
\begin{enumerate}
\item \textit{Energy constraints can be imposed over subgroups of nodes in the network:} We can have the reactive energy conservation constraint between subgroups of nodes, instead of on the network as a whole, i.e., $\sum \limits_{i=1}^{N_k} \Big(\lvert V_{ik} \rvert ^2 + \lvert I_{ik} \rvert ^2 \Big) = 1 \enskip \forall k=1,\ldots,M$. Here $M=$number of subgroups and $N_k=$ number of nodes in the $k^{\text{th}}$ subgroup. The update equations in this case are given by:
\begin{gather}\label{eq_prop1}
\dfrac{\partial V_{ik}(t)}{\partial t} =j\omega_k\sigma_{V_{ik}}(t)V_{ik}(t)-\Delta \sigma_{V_{ik}}(t)V_{ik}(t)\nonumber \\
\dfrac{\partial I_{ik}(t)}{\partial t} =j(\omega_k+\omega_{\phi_{ik}})\sigma_{I_{ik}}(t)I_{ik}(t)-\Delta \sigma_{I_{ik}}(t)I_{ik}(t) \nonumber \\
\tau_{ik} \omega_{\phi_{ik}}+\phi_{ik}(t) =g_{\phi_{ik}}(t) \quad \forall i,k,
\end{gather}
where $\omega_k$ is the constant angular frequency of the $k^{\text{th}}$ subgroup of nodes, and $\omega_{\phi_{ik}}=\dfrac{d\phi_{ik}(t)}{d t}$.
\item \textit{System dynamics and reactive-energy constraints remain invariant under the imposition of a global phase:} The network dynamics remain invariant to the imposition of a global phase component on all the network variables, and the conservation constraint is also satisfied in this case. The governing equations are given by:
\begin{align}
\label{eq_prop2}
\dfrac{\partial V_i(t)}{\partial t}& =j(\omega+\omega_{\phi_g})\sigma_{V_i}(t)V_{i}(t)-\Delta \sigma_{V_i}(t)V_i(t), \nonumber \\
\dfrac{\partial I_i(t)}{\partial t}& =j(\omega+\omega_{\phi_g}+\omega_{\phi_{i}})\sigma_{I_i}(t)I_{i}(t)-\Delta \sigma_{I_i}(t)I_i(t),
\end{align}
where $\phi_g$ is the global phase and $\omega_{\phi_g}=\dfrac{d \phi_g}{d t}$.
\item \textit{Reactive-energy constraints remain invariant with varying network dynamics under the imposition of a relative phase:} The conservation constraints are satisfied on the introduction of a relative phase component between the voltage and current phasors of each node, even though the overall network dynamics change. The governing equations are given by:
\begin{align}
\label{eq_prop3}
\dfrac{\partial V_i(t)}{\partial t}& =j(\omega+\omega_{\phi_{V_i}})\sigma_{V_i}(t)V_{i}(t)-\Delta \sigma_{V_i}(t)V_i(t), \nonumber \\
\dfrac{\partial I_i(t)}{\partial t}& =j(\omega+\omega_{\phi_{I_i}}+\omega_{\phi_{i}})\sigma_{I_i}(t)I_{i}(t)-\Delta \sigma_{I_i}(t)I_i(t),
\end{align} 
where $\phi_i=\phi_{I_i}-\phi_{V_i}$ is the relative external phase shift applied between the voltage and current phasors of the $i^{\text{th}}$ node, $\omega_{\phi_{V_i}}=\dfrac{d \phi_{V_i}}{d t}$ and $\omega_{\phi_{I_i}}=\dfrac{d \phi_{I_i}}{d t}$.
\item \textit{The model is dissipative and converges to limit cycle oscillations in steady state:} The second order time derivatives of Equations (\ref{eq_mainvol}) and (\ref{eq_maincurr}) lead to the following:
\begin{gather}
\dfrac{\partial^2 V_{i}}{\partial t^2}=j\omega\sigma_{V_i}\dot{V_{i}}+j\omega V_{i}\dot{\sigma_{V_i}}
-\Delta \sigma_{V_i}\dot{V_{i}}+\dot{\sigma_{V_i}}V_i\nonumber \\=\underbrace{-\omega^2\sigma_{V_i}^2V_{i}}_{\text{limit cycles}}+ \nonumber \\
\underbrace{[1+j\omega]\dot{\sigma}_{V_i}V_{i}-2j\omega \sigma_{V_i}\Delta \sigma_{V_i}V_i+(\Delta \sigma_{V_i})^2V_i}_{\text{dissipation}}, \label{eq_prop4V}\\
\dfrac{\partial^2 I_{i}}{\partial t^2}=j\omega^\prime_{\phi_i}\sigma_{I_i}\dot{I_{i}}+j\omega^\prime_{\phi_i} I_{i}\dot{\sigma_{I_i}}
-\Delta \sigma_{I_i}\dot{I_{i}}+\dot{\sigma_{I_i}}I_i\nonumber \\=\underbrace{-(\omega^\prime_{\phi_i})^2\sigma_{I_i}^2I_{i}}_{\text{limit cycles}} + \nonumber \\
\underbrace{[1+j\omega^\prime_{\phi_i}]\dot{\sigma}_{I_i}I_{i}-2j\omega^\prime_{\phi_i} \sigma_{I_i}\Delta \sigma_{I_i}I_i+(\Delta \sigma_{I_i})^2I_i}_{\text{dissipation}} \label{eq_prop4I},
\end{gather} 
 where $\omega^\prime_{\phi_i}=\omega+\omega_{\phi_i}$. The first terms in the RHS of Equations (\ref{eq_prop4V}) and (\ref{eq_prop4I}) correspond to stable limit cycle oscillations of all the phasors, while the other terms correspond to the dissipative effects in the network. This demonstrates that the network as a whole is essentially a coupled dissipative system that is capable of self-sustained oscillations under steady-state. Each individual state variable describing the network thus returns to the same position in its respective limit cycle at regular intervals of time, even when subjected to small perturbations.
\end{enumerate}

\begin{figure*}
\begin{center}
\includegraphics[page=1,scale=0.67]{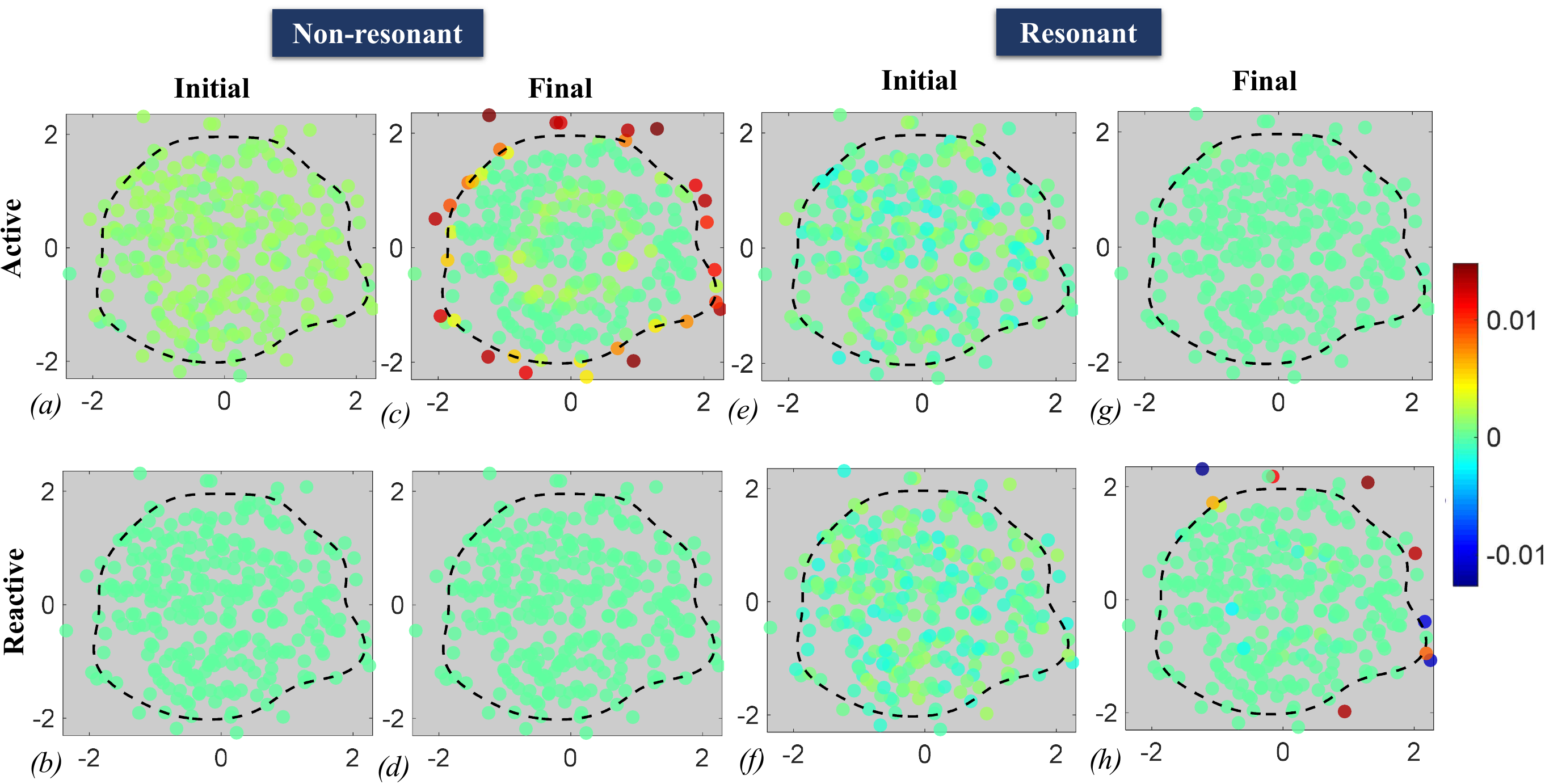}
\end{center}
\caption{Comparison of the active and reactive power dissipated at each node for the non-resonant model $\mathcal{M}_{nr}$ and the resonant model $\mathcal{M}_{r}$  for the synthetic one-class SVM problem on a two-dimensional dataset (Dataset I), with $N=300, \nu = 0.1, \omega=\pi/4$, and random initializations for $V_i,I_i,\phi_i \enskip \forall i=1,\ldots,N$: (a), (c): the values of the active power metric ($=\lvert V_{i} \rvert \lvert I_{i} \rvert $) at each node in the initial and final stages respectively for $\mathcal{M}_{nr}$; (b),(d): the values of the reactive power metric ($=0$) at each node in the initial and final stages respectively for $\mathcal{M}_{nr}$; (e), (g): the values of the active power metric ($=\lvert V_{i} \rvert \lvert I_{i} \rvert \cos \phi_i$) at each node in the initial and final stages respectively for $\mathcal{M}_{r}$; (f), (h): the values of the reactive power metric ($=\lvert V_{i} \rvert \lvert I_{i} \rvert \sin \phi_i$) at each node in the initial and final stages respectively for $\mathcal{M}_{r}$. For both models, $K(\cdot, \cdot)$ was chosen to be a Gaussian kernel with kernel parameter $\sigma=1$, and $\beta=1$ throughout the optimization process for $\mathcal{M}_{r}$.}
\label{fig_oneclass_actreact}
\end{figure*}

\section{Resonant Machine Learning Framework}\label{sec_learning}

In this section, we show how the framework introduced in Section \ref{sec_energymodel} can be applied for constructing resonant machine learning networks. In general, the framework can be applied to any learning network that optimizes a cost function defined over a set of learning variables $\alpha_i$ as 
\begin{gather}
\underset{\{\alpha_i\} } {\text{min}} \quad  \mathcal{H}(\{\alpha_i\})+h \Psi(\{\alpha_i\})  \nonumber \\
\textit{s.t.}\quad \sum_{i=1}^N \alpha_i = 1,\quad \alpha_i \ge 0 \quad \forall i=1,\ldots,N. \label{eq_probabilisticlearning1}
\end{gather} 
Here $\mathcal{H}(\{\alpha_i\})$ represents a loss-function~\cite{hastie2005elements} which depends on the learning problem (e.g. supervised, unsupervised or semi-supervised) and the dataset under consideration (e.g., training data). The second term $\Psi(\cdot)$ in the objective function is any linear or nonlinear function which represents either (a) a regularization function, or (b) a penalty function used to satisfy optimization constraints. $h$ is a hyperparameter which acts as a trade-off between $\mathcal{H}(\cdot)$ and $\Psi(\cdot)$. Because $\alpha_i$ could be viewed as probability measure, the optimization framework in Equation~(\ref{eq_probabilisticlearning1}) naturally lends itself to probabilistic learning models~\cite{murphy2012machine,ghahramani2015probabilistic, chakrabartty2007gini}.

\par The above problem can be  mapped to the resonant learning framework in Section \ref{sec_energymodel} by substituting $\alpha_i=\lvert V_i \rvert ^2+\lvert I_i \rvert ^2$, to arrive at the following problem:
\begin{gather}
\underset{\{\lvert V_i \rvert,\lvert I_i \rvert,\phi_i\} } {\text{min}} \quad  \mathcal{H}(\{\lvert V_i \rvert,\lvert I_i \rvert\})+h \Psi(\{\lvert V_i \rvert,\lvert I_i \rvert\})\nonumber \\
+ \beta\sum\limits_{i=1}^N \lvert V_i\rvert^2\lvert I_i\rvert^2\cos^2\phi_i  \nonumber \\
\textit{s.t.}\quad \sum_{i=1}^N(\lvert V_i \rvert ^2+\lvert I_i \rvert ^2) = 1, \quad \lvert \phi \rvert \le \pi \quad \forall i=1,\ldots,N  \label{eq_probabilisticlearning2}
\end{gather}
Note that non-probabilistic learning problems can also be mapped to the probabilistic framework by imposing an additional constraint, as discussed in Appendix A.

\subsection{One-class resonant SVM}\label{subsubsec_dissipative}

We now show how the framework in Equation~(\ref{eq_probabilisticlearning2}) can be used to design a resonant one-class SVM.
\par The solution of a generic one-class SVM is obtained by solving the following optimization problem \cite{scholkopf2000support,tax1999support,gangopadhyay2017extended}:
\begin{gather}
\underset{\{\alpha_i\} } {\text{min}} \quad  \dfrac{1}{2} \sum\limits_{i=1}^N\sum \limits_{j=1}^N \alpha_iK(\x_i,\x_j)\alpha_j  \label{eq_oneclass1}\nonumber \\
\textit{s.t.}\quad \sum_{i=1}^N \alpha_i = 1,\quad 0<\alpha_i < \dfrac{1}{\nu N} \quad \forall i=1,\ldots,N  \label{eq_oneclassconstraint}
\end{gather}
where $\mathbf{X}=[\mathbf{x_1},\ldots,\mathbf{x_i},\ldots,\mathbf{x_N}]\in \mathbb{R}^{N \times D}$ is the $D-$dimensional input dataset of size $N$, $\nu \in \{0,1\}$ is a parameter which controls the size of the decision surface, $K(\cdot,\cdot)$ is a positive definite Kernel function satisfying Mercer's conditions, and $\alpha_i$'s are the learning parameters. The optimization problem above can be reformulated by replacing the inequality constraint with a smooth penalty or loss function $\Psi(\cdot)$ like the logarithmic barrier, e.g., $\Psi(\alpha_i,\nu)=-\log(\dfrac{1}{\nu N}-\alpha_i)$. 
\begin{gather}
\underset{\{\alpha_i\}} {\text{min}} \quad  \mathcal{H}=\dfrac{1}{2} \sum\limits_{i=1}^N\sum \limits_{j=1}^N \alpha_iK(\x_i,\x_j)\alpha_j \ +h\sum \limits_{i=1}^N \Psi(\alpha_i,\nu)\nonumber \\
\textit{s.t.}\quad \sum_{i=1}^N \alpha_i = 1,
\end{gather}
The parameter $h$ determines the steepness of the penalty function, where a lower value of $h$ implies an almost-accurate inequality constraint. 

\begin{figure*}
\begin{center}
\includegraphics[page=1,scale=0.13,trim=8 6 6 6,clip]{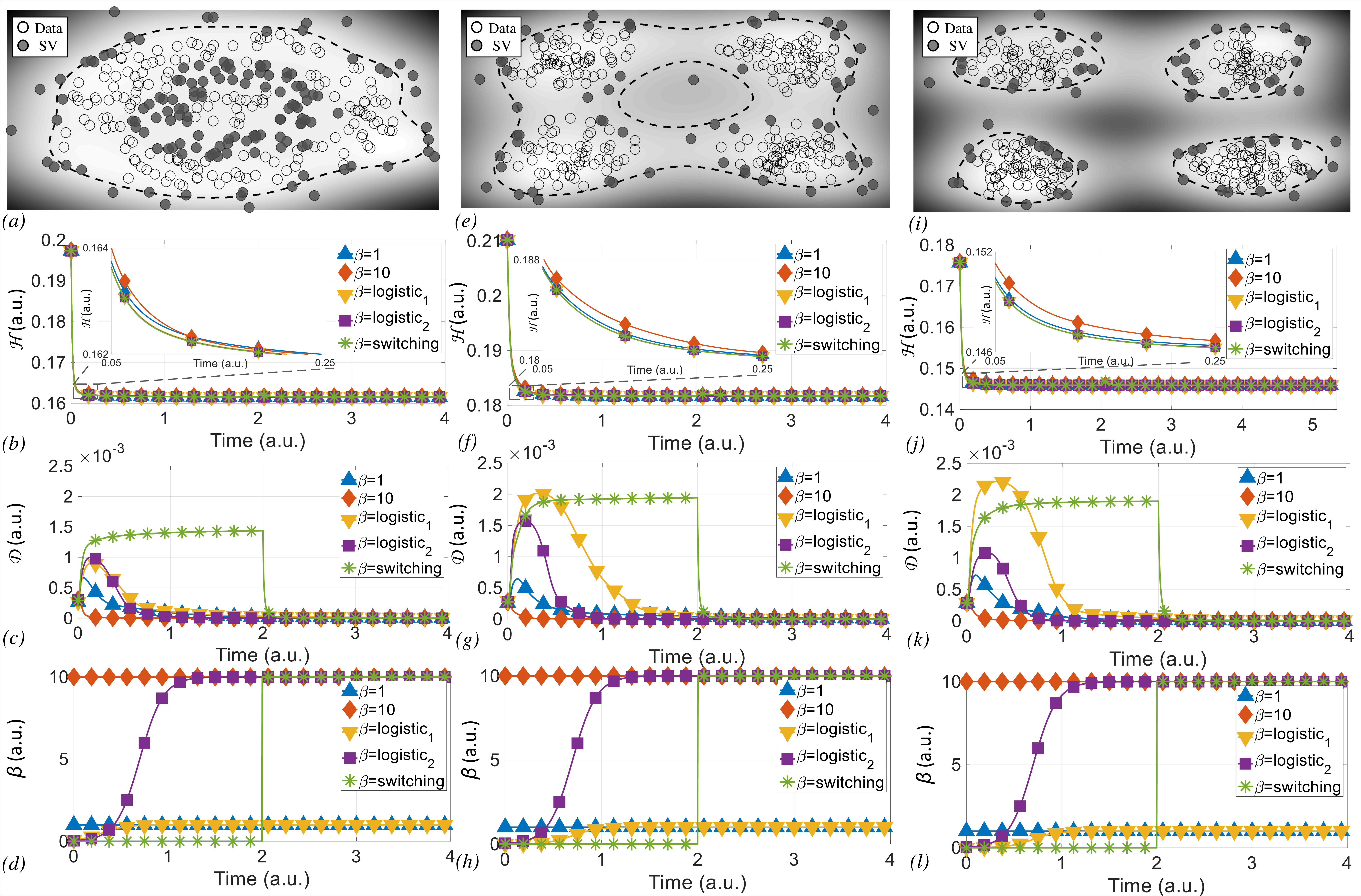}
\end{center}
\caption{Comparison of the performance of the resonant model $\mathcal{M}_{r}$ for different choices of annealing schedules for $\beta$ for the one-class SVM problem, on three different two-dimensional synthetic datasets for a simulation duration of $t = 4  \enskip a.u.$. In all cases, $N=300, \nu = 0.1, \omega=\pi/20$, and $K(\cdot,\cdot)$ was chosen to be a Gaussian kernel with parameter values $\sigma=1,10$ and $20$ for synthetic Dataset I, II and III respectively. Additionally for each dataset, $V_i,I_i,\phi_i$ were randomly initialized $\forall i=1,\ldots,N$: (a) contour plot with the decision boundary around the datapoints and support vectors (SVs); (b) time evolution of $\mathcal{H}$ (inset shows a zoomed in view of the cost evolution in the transient phase); (c) time evolution of $\mathcal{D}$ and (d) time evolution of $\beta$ for different annealing schedules. The curves corresponding to $\beta=1$ and $10$ denote the cases when $\beta$ takes a constant value throughout the simulation duration; $\beta=\text{logistic}_1$ and $\beta=\text{logistic}_2$ correspond to the cases when $\beta$ is slowly increased following a logistic curve, and takes on maximum values of $\beta_{\text{max}}=1$ and $\beta_{\text{max}}=10$ respectively; $\beta=\text{switching}$ corresponds to the case when  $\beta$ switches from a minimum value($\beta_{\text{min}}=0$) to a maximum value($\beta_{\text{max}}=10$) at $t=2  \enskip a.u.$, after the system has converged to the optimal solution. (e)-(h) show similar plots on Dataset II, while (i)-(l) show the plots corresponding to Dataset III.}
\label{fig_comparison}
\end{figure*} 
 
\begin{figure}
\begin{center}
\includegraphics[page=1,scale=0.58]{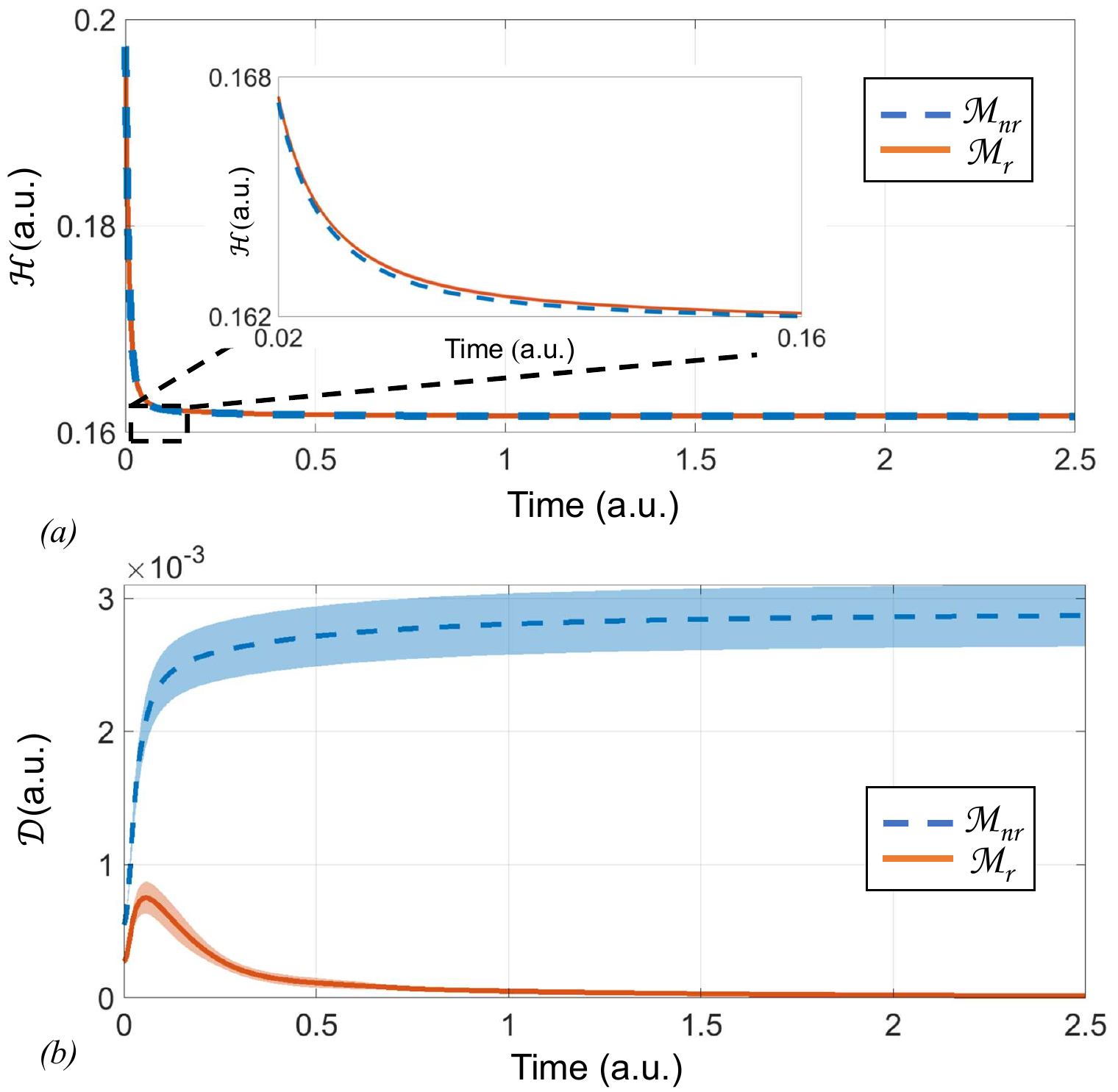}
\end{center}
\caption{Robustness to random initialization: Comparison of (a) the time-evolution of the cost $\mathcal{H}$ and (b) the dissipation metric profile $\mathcal{D}$  ($\sum \limits_{n=1}^N \lvert V_i \rvert^2  \lvert I_i \rvert^2$ for the non-resonant model $\mathcal{M}_{nr}$ and $\sum \limits_{n=1}^N \lvert V_i \rvert^2  \lvert I_i \rvert^2 \cos^2 \phi_i$ for the resonant model $\mathcal{M}_{r}$ respectively) for the synthetic one-class SVM problem on Dataset I for $N=300, \nu = 0.1,\omega=\pi/8$ over 10 random initializations for $V_i,I_i,\phi_i \enskip \forall i=1,\ldots,N$. The regularization parameter was chosen to be $\beta=1$ for the entire simulation duration of $t = 4  \enskip a.u.$ of the optimization process for $\mathcal{M}_{r}$. For all the curves, the solid line indicates the mean response, while the shaded region shows the standard deviation about the mean across the trials.}
\label{fig_oneclass_costdissp}
\end{figure}

\par An equivalent complex-domain representation in terms of voltages and currents in an LC network can be arrived at if we consider $\alpha_i=\lvert V_i \rvert ^2+\lvert I_i \rvert ^2 \enskip \forall i$. In this case, we consider that the network is globally energy constrained, and all the individual units in the network have the same frequency $\omega$. The redefined learning problem is as follows:
\begin{gather}
\label{eq_dissipativecost2}
\underset{\{\lvert V_i \rvert,\lvert I_i \rvert\}} {\text{min}} \quad  \mathcal{H}=\dfrac{1}{2} \sum\limits_{i=1}^N\sum \limits_{j=1}^N (\lvert V_i \rvert ^2+\lvert I_i \rvert ^2)K(\x_i,\x_j)\nonumber \\
 \times (\lvert V_j \rvert ^2+\lvert I_j \rvert ^2) +h\sum \limits_{i=1}^N \Psi(\lvert V_i \rvert,\lvert I_i \rvert ,\nu)\nonumber \\
\textit{s.t.}\quad \sum_{i=1}^N (\lvert V_i \rvert ^2+\lvert I_i \rvert ^2) = 1.
\end{gather}

Introducing the active-power dissipation regularization we arrive at the following problem: 
\begin{gather}\label{eq_nondissipativecost}
\underset{\{\lvert V_i \rvert,\lvert I_i \rvert,\phi_i\}} {\text{min}} \quad  \mathcal{L}=\dfrac{1}{2} \sum\limits_{i=1}^N\sum \limits_{j=1}^N (\lvert V_i \rvert ^2+\lvert I_i \rvert ^2)K(\x_i,\x_j)
 \nonumber \\
\times (\lvert V_j \rvert ^2+\lvert I_j \rvert ^2)  + h\sum \limits_{i=1}^N \Psi(\lvert V_i \rvert ,\lvert I_i \rvert,\nu) \nonumber \\
+ \beta\sum\limits_{i=1}^N \lvert V_i\rvert^2\lvert I_i\rvert^2\cos^2\phi_i \nonumber \\
\textit{s.t.}\quad \sum_{i=1}^N (\lvert V_i \rvert ^2+\lvert I_i \rvert ^2) = 1, \quad \lvert \phi_i \rvert \le \pi \enskip \forall i.
\end{gather}
\par The update equations in this case are of the form shown in Equations (\ref{eq_mainvol})-(\ref{eq_mainphase}). Figure \ref{fig_oneclass_actreact} shows a comparison of the active and reactive power metrics of each node of the non-resonant model $\mathcal{M}_{nr}$  and the resonant model $\mathcal{M}_{r}$  for a synthetic one-class SVM problem on a two-dimensional dataset (Dataset I). The dataset was generated by uniformly selecting $300$ random points within a circle having a fixed radius. Here $N=300, \nu = 0.1, \omega=\pi/4$, with random initializations for $V_i,I_i,\phi_i \enskip \forall i=1,\ldots,N$. A constant value of the regularization hyperparameter $\beta=1$ was considered throughout the duration of the optimization process for $\mathcal{M}_{r}$. Figures \ref{fig_oneclass_actreact}(a) and (c) show the values of the active power metric ($=\lvert V_{i} \rvert \lvert I_{i} \rvert $) at each node in the initial and final stages respectively for $\mathcal{M}_{nr}$, while Figures \ref{fig_oneclass_actreact}(b) and (d) show the values of the reactive power metric ($=0$) at each node in the initial and final stages respectively for $\mathcal{M}_{nr}$. Similarly, Figures \ref{fig_oneclass_actreact}(e) and (g) show the values of the active power metric ($=\lvert V_{i} \rvert \lvert I_{i} \rvert \cos \phi_i$) at each node in the initial and final stages respectively for $\mathcal{M}_{r}$ while Figures \ref{fig_oneclass_actreact}(f) and (h), finally, show the values of the reactive power metric ($=\lvert V_{i} \rvert \lvert I_{i} \rvert \sin \phi_i$) at each node in the initial and final stages respectively for $\mathcal{M}_{r}$.


\par Figure \ref{fig_comparison} shows a comparison of the performance of the resonant model $\mathcal{M}_{r}$ for different choices of annealing schedules for $\beta$ on three different two-dimensional synthetic datasets for a simulation duration of $t = 4  \enskip a.u.$. In all cases, $N=300, \nu = 0.1, \omega=\pi/20$, and $K(\cdot,\cdot)$ was chosen to be a Gaussian kernel with parameter values $\sigma=1,10$ and $20$ respectively for synthetic Dataset I, II and III. Additionally for each dataset, $V_i,I_i,\phi_i$ were randomly initialized $\forall i=1,\ldots,N$. Figure \ref{fig_comparison}(a) shows the contour plot with the decision boundary around the datapoints, along with the support vectors (SVs) for Dataset I, while Figures \ref{fig_comparison}(b), (c) and (d) show the time evolutions of the cost $\mathcal{H}$, the dissipation metric $\mathcal{D}$ and the hyperparameter $\beta$ for different annealing schedules. The curves corresponding to $\beta=1$ and $10$ denote the cases when $\beta$ takes a constant value throughout the simulation duration. $\beta=\text{logistic}_1$ and $\beta=\text{logistic}_2$ correspond to the cases when $\beta$ is slowly increased following a logistic curve of the form $\beta(t)=\beta_{\text{min}}+\dfrac{\beta_{\text{max}}-\beta_{\text{min}}}{1+\exp(-k(t+t_0))}$, and takes on maximum values of $\beta_{\text{max}}=1$ and $\beta_{\text{max}}=10$ respectively starting from $\beta_{\text{min}}=0$. Finally, $\beta=\text{switching}$ corresponds to the case when  $\beta$ switches from a minimum value ($\beta_{\text{min}}=0$) to a maximum value ($\beta_{\text{max}}=10$) at $t=2  \enskip a.u.$, after the system has converged to the optimal solution. Figures \ref{fig_comparison}(e)-(h) show similar plots on Dataset II, while Figures \ref{fig_comparison}(i)-(l) show the plots corresponding to Dataset III. Dataset I, as described before, was generated by selecting $300$ points uniformly within a two-dimensional circle having a fixed radius. Datasets II and III also consist of $300$ data points generated using a Gaussian mixture model consisting of four different clusters, with a fixed cluster mean and variance associated with each cluster. Dataset II and III differ only in terms of the cluster means associated with their respective constituent clusters, while the cluster variances are the same for all the clusters for both the datasets. It can be seen that since the optimization problem is convex, the model always converges to the optimal solution for every dataset, irrespective of the the annealing schedule for $\beta$ or the dataset complexity. However, the dissipation profiles corresponding to a particular annealing schedule strongly depend on the complexity of the dataset. In general, however, higher values of $\beta$ would lead to lower dissipation profiles during the learning process. Also, the model shows a much slower convergence in terms of the actual objective function for a constant non-zero value of $\beta$ throughout the optimization compared to the case when $\beta$ is slowly annealed to a sufficiently high value. The choice of a proper annealing schedule for $\beta$ would thus involve a trade-off between the speed of convergence and the rate of power dissipation.
\par Figures \ref{fig_oneclass_costdissp}(a) and (b) finally, show the robustness of the proposed framework to different initial conditions by providing a comparison of the time evolutions of the cost $\mathcal{H}$ and the dissipation metric $\mathcal{D}$ ($\sum \limits_{n=1}^N \lvert V_i \rvert^2  \lvert I_i \rvert^2$ for the non-resonant model $\mathcal{M}_{nr}$ and $\sum \limits_{n=1}^N \lvert V_i \rvert^2 \lvert I_i \rvert^2 \cos^2 \phi_i$ for the resonant model $\mathcal{M}_{r}$ respectively), when applied to Dataset I. We show the results for 10 random initializations of $V_i,I_i$ and $\phi_i, \enskip \forall i$. In all cases, $\nu = 0.1, \omega=\pi/8$, and $K(\cdot,\cdot)$ was chosen to be a Gaussian kernel with parameter value $\sigma=1$. Note that even though the 10 trails had different initializations of $V_i,I_i$, they were chosen to have the same initial value of $\alpha_i'$s in all cases, because of which there is no deviation between the cost evolution curves for both $\mathcal{M}_{nr}$ and $\mathcal{M}_{r}$. The dissipation evolution is however different across different trials for both the models. However, the dissipation attains a final value of zero for $\mathcal{M}_{r}$ for all the trials, while there is a finite dissipation for $\mathcal{M}_{nr}$ in all cases.

\begin{table}[!htbp] 
	\centering
		\begin{minipage}{\linewidth}
			\renewcommand{\thetable}{\arabic{table}}
			\caption{Performance on real benchmark datasets for $\nu = 0.1$}
			\label{tab_realdata}
			\renewcommand{\arraystretch}{2}
			\centering
\begin{tabular}{|c|c|c|c|c|}\hline
	{Dataset} & {Majority Class Size} & {Correct} & {Outliers} & {SVs} \\ \hline
	Iris & (50,4) & 48 & 2 & 8 \\
 \hline
	Heart  &  (139,13) & 1210 & 18 & 56  \\ \hline
	Diabetes  &  (500,4) & 466 & 34 & 90 \\
	\hline 
	Ecoli &  (327,7) & 309 & 18 & 42 \\
	\hline 
Adulta3a  &  (2412,123) & 2111 & 301 & 356 \\
	\hline 
Mammography  &  (10923,6) & 9608 & 1315 & 5969 \\
\hline 
\end{tabular}
	\end{minipage}
\end{table}

\begin{figure*}
	\begin{center}
		\includegraphics[page=1,scale=0.13,trim=8 8 8 8,clip]{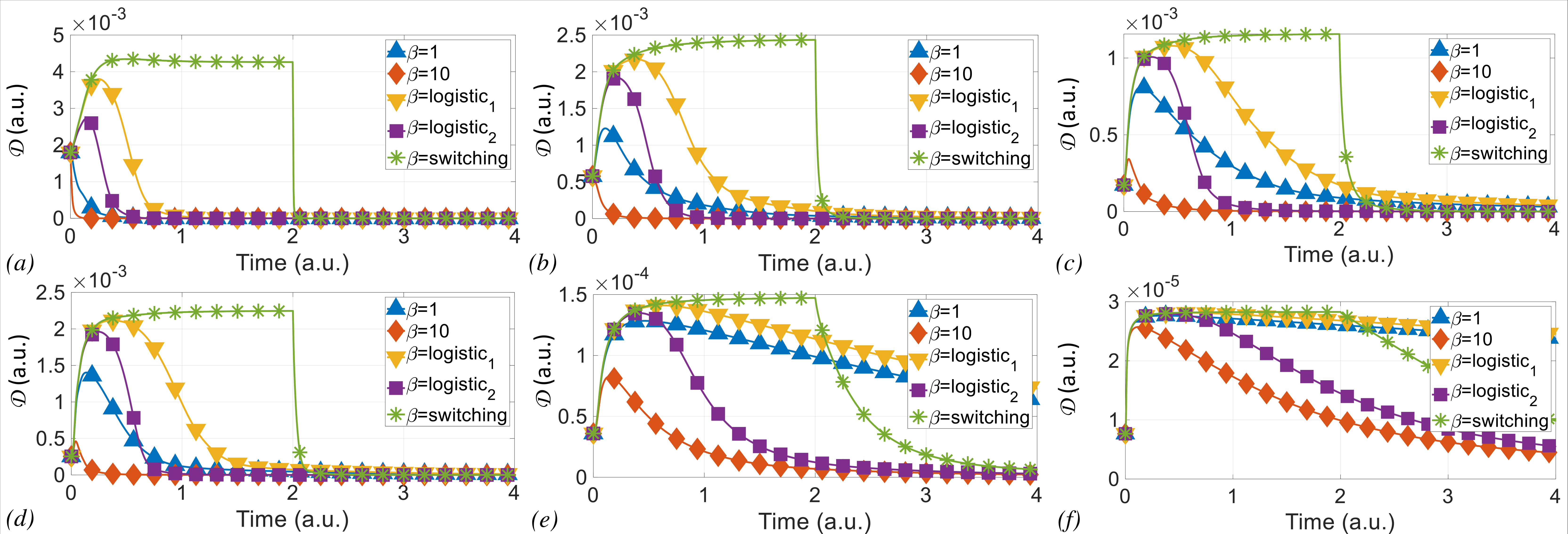}
	\end{center}
	\caption{Comparison of the dissipation profiles of the resonant model $\mathcal{M}_{r}$ for different choices of annealing schedules for $\beta$ for different real-life benchmark datasets :(a) `Iris'; (b) `Heart'; (c) `Diabetes; (d) `Ecoli'; (e) `Adulta3a'; and (f) `Mammography'.}
	\label{fig_comparison_realdata}
\end{figure*}

\par We also conducted experiments on real-life benchmark datasets of varying sizes and dimensionality, considering the majority class as inliers in all cases. In all the experiments, we used similar parameter settings and annealing schedules as used in the experiments shown in Figure \ref{fig_comparison}, and $\nu=0.1$. Table \ref{tab_realdata} shows the dataset description along with the performance (number of inliers, number of outliers and number of support vectors) of the one-class SVM classifier for the following datasets: `Iris' (inlier: class `setosa'), `Heart' (inlier: `healthy' heart), `Diabetes' (inlier: non-diabetic patients), `Ecoli' (inlier: classes `cp', `im', `pp', `imU' and `om'), `Adulta3a' (inlier: individuals with income$\le \$50K$) and `Mammography' (inlier: non-calcifications)\cite{Dua2019,Rayana2016}.  Figures \ref{fig_comparison_realdata} (a)-(f) show the dissipation profiles corresponding to different annealing schedules for the different datasets. It can be seen that an optimal choice of the annealing schedule depends on both the dataset size and complexity, even though the dissipation decreases over time for all the cases, irrespective of the schedule chosen.


\par
\begin{figure}
\begin{center}
\includegraphics[page=1,scale=0.65]{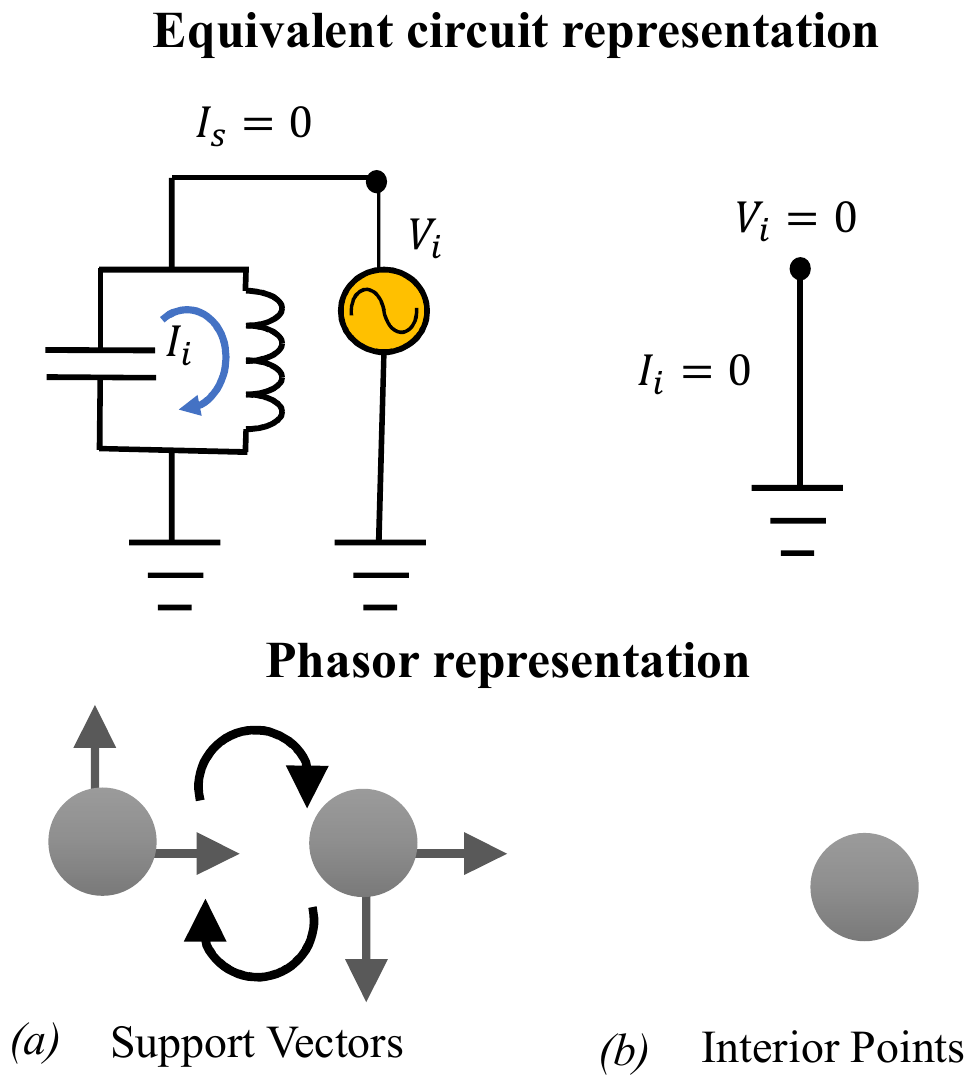}
\end{center}
\caption{Circuit based and phasor based representations for a one-class SVM problem: (a) support vectors, corresponding to resonant LC tanks and (b) interior points, corresponding to sinks/ground ($V_i,I_i=0$).  }
\label{fig_svmvisualize}
\end{figure}

 Interestingly, the solution of the one-class SVM can be interpreted in terms of an electrical network as follows, as shown in Figure \ref{fig_svmvisualize}: (a) the support vectors have voltage and current magnitudes with a $\pm \pi/2$ phase shift between them, hence can be interpreted as resonant LC tanks; and
(b)	the interior points well inside the boundary have both zero voltage and current magnitudes, and can be essentially treated as floating sinks.

\section{Discussions and Conclusions}
\label{sec_discussions}
 
In this paper we proposed a complex-domain formulation of a machine learning network that ensures that 
the network's active power is dissipated only during the process of learning whereas the network's reactive power is maintained to be zero at all times. We showed that the active power dissipation during learning can be controlled using a
phase regularization parameter. Also, the framework is robust to variations in the initial conditions and to the choice of the input/driving frequency $\omega$. The proposed approach thus provides a physical
interpretation of machine learning algorithms, where the energy required for storing learned parameters is sustained
by electrical resonances due to nodal inductances and nodal capacitances. Using the one-class support vector machine problem as a case study, we have shown how the steady state solution of a learning problem can be interpreted in terms of these nodal circuit elements. 


\par Future directions involve exploring the implications of incorporating network dynamics in the free frequency variable $\omega$, and utilizing the phase information associated with each node in the learning process. Also, the experimental results presented in this paper were based on an unsupervised learning setting. Exploring an energy-efficient framework for supervised learning problems would also be a part of our future work.

\par In this paper we also proposed a dynamical system model based on complex-domain growth transforms. The formulation
is general enough to be applicable to other complex-domain learning models\cite{hirose2003complex,trabelsi2017deep,gaudet2018deep,guberman2016complex,bouboulis2014complex}. Our proposed framework also preserves both the magnitude and phase information, and provides additional flexibilty compared to other complex domain learning models in terms of phase manipulation/cancellation \cite{trabelsi2017deep}. 

\par In addition to implementing classic machine learning algorithms, the complex growth transform dynamical system can also be used for designing synchronized networks of coupled oscillators. Such networks can be potentially used for solving different computing tasks like optimization, pattern matching etc. as is achievable using coupled oscillatory computing models\cite{raychowdhury2019computing,torrejon2017neuromorphic}. An oscillator network designed in this fashion is capable of demonstrating stable, self-sustaining oscillatory dynamics, whereby the network can return to its initial stable limit cycle configuration following small perturbations, while simultaneously minimizing some underlying system-level objective function. The framework could also be used to study connections between learning and synchronization, or the emergence of a rhythmic periodic pattern exhibited by a group of coupled oscillators, which provides the key to understanding periodic processes pervading complex networks of different biological, physical, social and quantum ensembles \cite{strogatz2000kuramoto,nielsen2010quantum}. In this regard, the existing mathematical models for such collective behavior are mostly phenomenological or bottom-up, and in general do not provide a network-level perspective of the underlying physical process. The proposed growth-transform formulation, thus, could provide new network-level insights
into the emergence of phase synchronization, phase cohesiveness and frequency synchronization in coupled-oscillator networks.   

\par Note here that since we are implicitly assuming an analog implementation, the learning network should converge to the steady-state solution of the optimization, where the time of convergence depends on the network’s time-constant \cite{gangopadhyay2018continuous}. Hence the notion of time complexity is not well-defined. However, in a digital implementation, the time complexity would depend on the learning algorithm under consideration, dataset size and dimensionality, angular frequency $\omega$, and time constants $\tau_i$ for the phase updates.

\section{Appendix}

\subsection{Resonance in an LC tank}

\begin{wrapfigure}{l}{0.20\textwidth}
	\centering
		\includegraphics[page=1,scale= 0.01, width=0.19\textwidth, clip, trim = 2 2 2 2]{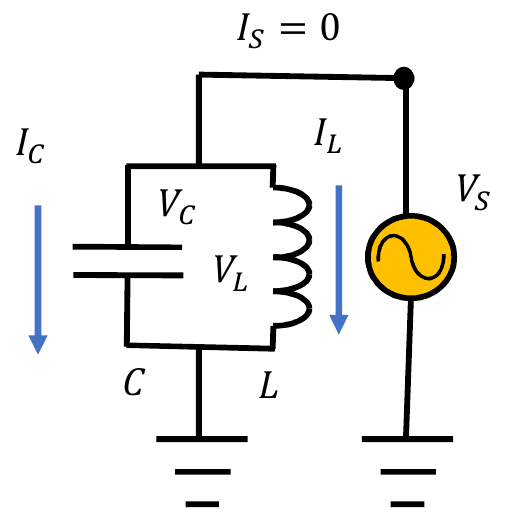}
	\caption{LC tank resonator}
	\label{fig_lctank}
\end{wrapfigure}

Consider the parallel LC tank circuit shown in Figure \ref{fig_lctank}, with $V_C$ and $V_L$ being the voltages across the capacitor $C$ and inductor $L$ respectively. $I_C$ and $I_L$ denote the corresponding currents flowing through the elements. Thus, $V_S=V_L=V_C$ and $I_S=I_L+I_C$.
Considering the LC tank to be driven by the  voltage source $V_S$ at frequency $\omega$, we have the following condition in steady state:
\begin{equation}
I_S(\omega)=\dfrac{V_S(\omega)}{j\omega L}[1-\omega^2LC]
\end{equation}
Resonant condition of the circuit is achieved when
\begin{align}
\omega =\dfrac{1}{\sqrt{LC}} 
\implies I_S(\omega)=0
\end{align}
This implies that the apparent power $S_N=P_N+jQ_N=V_SI_S^*+V_LI_L^*+V_CI_C^*$, where the active power $P_N=0$. Additionally at resonance, the reactive power $Q_N=Q_C+Q_L=V_LI_L^*+V_CI_C^*=-\dfrac{j}{\omega L}\lvert V(\omega)\rvert ^2+\dfrac{j}{\omega L}\lvert V(\omega)\rvert ^2=0$. Here $Q_C$ and $Q_L$ are the reactive powers associated with the capacitance and inductance respectively. 

\subsection{Mapping a generic optimization problem to the equivalent network model}
\label{subsec_map}
\par  Let us consider an optimization problem defined over a probabilistic domain, given by the following generic form:

\begin{align}
\underset{\{x_i\}} {\text{min}} \quad  \mathcal{H}(\{x_i\})  \nonumber  \\
\textit{s.t.}\quad \sum_{i=1}^N  x_i=1, \enskip x_i\ge 0 \label{eq_actualopt}
\end{align}
\par We can map the above to the electrical network-based model introduced in Section \ref{sec_energymodel} by replacing $x_i=\lvert V_i \rvert^2+\lvert I_i \rvert^2$, which leads to the following problem in the $\{\lvert V_i \rvert^2,\lvert I_i \rvert^2\}$ domain:
\begin{align}
\underset{\{\lvert V_i \rvert,\lvert I_i \rvert\}} {\text{min}} \quad  \mathcal{H}(\{\lvert V_i \rvert,\lvert I_i \rvert\})  \nonumber  \\
\textit{s.t.}\quad \sum_{i=1}^N \Big( \lvert V_i \rvert^2+\lvert I_i \rvert^2\Big)=1. 
\end{align}
Note that the method also works for optimization problems defined over non-probabilistic domains, of the following form:
\begin{align}
\underset{\{x_i\}} {\text{min}} \quad  \mathcal{H}(\{x_i\})  \nonumber  \\
\textit{s.t.} \enskip \lvert x_i \rvert \le 1, \enskip x_i \in \mathbb{R} \enskip \forall i=1, \ldots, N. 
\end{align}
This can be done by considering $x_i=x_i^+-x_i^- \enskip \forall i$, where both $x_i^+,x_i^- \ge 0$. Since by triangle inequality, $\lvert x_i \rvert \le \lvert x_i^+ \rvert + \lvert x_i^- \rvert$, enforcing $x_i^++x_i^-=1 \enskip \forall i$ would automatically ensure $\lvert x_i \rvert \le 1 \enskip \forall i$. Thus we have,
\begin{align}
\underset{\{x_i\}} {\text{argmin}} \quad \mathcal{H}(\{x_i\})&  \equiv \underset{\{x_i^+,x_i^-\}} {\text{argmin}} \quad \mathcal{H}(\{x_i^+,x_i^-\}) \label{eq_twovar}\\
 \textit{s.t.} \quad \lvert x_i \rvert \le 1, \enskip x_i \in \mathbb{R} & \quad \quad  \textit{s.t.} \quad x_i^++x_i^-=1, \enskip x_i^+,x_i^- \ge 0  \nonumber 
\end{align}
\par In this case, we replace $x_i^+=\lvert V_i \rvert^2, \enskip x_i^-=\lvert I_i \rvert^2$, and the equivalent problem in the $\{\lvert V_i \rvert^2,\lvert I_i \rvert^2\}$ domain is thus as follows:
\begin{align}
\underset{\{\lvert V_i \rvert,\lvert I_i \rvert\}} {\text{min}} \quad  \mathcal{H}(\{\lvert V_i \rvert,\lvert I_i \rvert\})  \nonumber  \\
\textit{s.t.} \enskip\lvert V_i \rvert^2+\lvert I_i \rvert^2=1,  \enskip \forall i=1, \ldots, N, \label{eq_optform2}
\end{align}
\par For example, the variables $\{x_i\}$ can represent the Lagrangian multipliers in the primal space of a support vector machine network, or the weights and biases of a generic neural network.

\subsection{Complex growth transform dynamical system (Table \ref{algo1})}
\label{subsec_thm1}
Let us consider the optimization problem in Equation (\ref{eq_actualopt}) again. We can use the Baum-Eagon inequality \cite{baum1968growth,gopalakrishnan1989generalization} to converge to the optimal point of $\mathcal{H}$ in steady state, by using updates of the form
\begin{equation}
x_i \leftarrow \dfrac{x_i\Big(-\dfrac{\partial \mathcal{H}(\{x_i\})}{\partial x_i}+\lambda \Big)}{\sum \limits_{k=1}^N x_k\Big(-\dfrac{\partial \mathcal{H}(\{x_k\})}{\partial x_k}+\lambda \Big)}, \label{eq_growthdyn1}
\end{equation} 
Here, $\mathcal{H}$ is assumed to be Lipschitz continuous \cite{chatterjee2018decentralized} on the domain $D=\{x_1,\ldots,x_N :\sum\limits_{i=1}^{N} x_i=1, \enskip x_i \ge 0 \enskip \forall i \} \subset \mathbb{R}^N_+$. The constant $\lambda \in \mathbb{R}_+$ is chosen such that 
$\Big \lvert -\dfrac{\partial \mathcal{H}(\{x_i\})}{\partial x_i}+\lambda\Big \rvert > 0, \forall i$.

We can solve the optimization problem given by Equation (\ref{eq_optimclass}) by using the growth transforms discussed above. The outline of the proof is as follows: (1) We will start with a generic magnitude domain optimization problem without any phase regularizer and derive the form for the growth transform dynamical system which would converge to the optimal point asymptotically; (2) Next, we derive a complex domain counterpart of the above, again without phase constraints; (3) Finally, we derive the complex domain dynamical system by incorporating a phase regularizer in the objective function.
\par Since the time evolutions of $V_i$ and $I_i$ are symmetric because of the conservation constraints, for the rest of the section we will consider only the update equations for the voltages $V_i$, and similar results would also apply to the updates for $I_i$.

\subsubsection{\textbf{Condition 1}}
\par Considering $\beta=0$ in Equation (\ref{eq_optimclass}) and $\mathcal{H}(\{\lvert V_i \rvert,\lvert I_i \rvert\})$ to be Lipschitz continuous over the domain $D=\{\lvert V_i \rvert,\lvert I_i \rvert :\sum\limits_{i=1}^{N}\Big( \lvert V_i\rvert^2 + \lvert I_i\rvert^2 \Big)=1\}$, we can use the growth transforms to arrive at the following discrete-time update equations in terms of the voltage and current magnitudes :

\begin{align}
\lvert V_{i,n}\rvert^2\leftarrow & g_{V_i,n-1}(\{\lvert V_{i,n-1} \rvert^2,\lvert I_{i,n-1} \rvert^2\}) ,
\end{align}
where
\begin{gather}
g_{V_i,n-1}(\{\lvert V_{i,n-1} \rvert^2,\lvert I_{i,n-1} \rvert^2\})=\dfrac{\lvert V_{i,n-1}\rvert^2}{\mu_{n-1}}\Big(-\dfrac{\partial \mathcal{H}}{\partial \lvert V_{i,n-1}\rvert^2}+\lambda\Big), \\
\mu_{n-1}=\sum\limits_{k=1}^N \Big(\lvert V_{k,n-1}\rvert^2\Big[-\dfrac{\partial \mathcal{H}}{\partial \lvert V_{k,n-1}\rvert^2}+\lambda \Big] \nonumber \\
+\lvert I_{k,n-1}\rvert^2\Big[-\dfrac{\partial \mathcal{H}}{\partial \lvert I_{k,n-1}\rvert^2}+\lambda \Big]\Big),
\end{gather} 
 and $\lambda \in \mathbb{R}_+$ is chosen to ensure  
that $\Big(-\dfrac{\partial \mathcal{H}}{\partial \lvert V_i\rvert^2}+\lambda \Big)>0$ and $\Big( -\dfrac{\partial \mathcal{H}}{\partial \lvert I_i\rvert^2}+\lambda \Big)> 0, \forall i$. 
 
\par Writing $g_{V_i,n-1}=g_{V_i,n-1}(\{\lvert V_{i,n-1} \rvert^2,\lvert I_{i,n-1} \rvert^2\})$ for notational convenience and taking $g_{V_i,n-1}=\lvert V_{i,n-1}\rvert^2 \sigma_{V_i,n-1}^2$, we get:
\begin{align}
 \lvert V_{i,n}\rvert^2 
&\leftarrow \lvert V_{i,n-1}\rvert^2\sigma_{{V_i},n-1}^2 
\end{align}
\subsubsection{\textbf{Condition 2}}

\textit{Considering $\beta=0$ in Equation (\ref{eq_optimclass}) and $V_i,I_i\in D^C=\{V_i,I_i\in \mathbb{C} :\sum\limits_{i=1}^N \Big( \lvert V_i\rvert^2 + \lvert I_i\rvert^2\Big)=1\}$, a time evolution of the form given below converges to limit cycles corresponding to the optimal point of a Lipschitz continuous objective function $\mathcal{H}(\{\lvert V_i \rvert,\lvert I_i \rvert\})$:}
\begin{align}
\label{eq_thm1}
{V_{i,n}} & \leftarrow V_{i,n-1}\sigma_{{V_i},n-1} e^{j\theta} ,\nonumber \\
{I_{i,n}} & \leftarrow I_{i,n-1}\sigma_{{I_i},n-1} e^{j(\theta+\phi_i)}
\end{align}
\textit{where $\sigma_{V_i},\sigma_{I_i} \rightarrow 1 \quad \forall i=1,\ldots,N$, in steady state, $\theta$ is the instantaneous global phase difference of the system of phasors with respect to an absolute stationary reference frame, and $\phi_i$ is the instantaneous phase difference between $V_i$ and $I_i$.}
\newline{Proof:}

Since $\lvert V_{i,n} \rvert^2=V_{i,n}V_{i,n}^*$, $\lvert I_{i,n} \rvert^2=I_{i,n}I_{i,n}^*$, the update equations can be rewritten as:
\begin{align}
V_{i,n}V_{i,n}^* & \leftarrow [V_{i,n-1}\sigma_{{V_i},n-1} e^{j\theta}]
\times [V_{i,n-1}\sigma_{{V_i},n-1} e^{j\theta}]^* 
\nonumber \\
I_{i,n}I_{i,n}^* & \leftarrow [I_{i,n-1}\sigma_{{I_i},n-1} e^{j(\theta+\phi_i)}]
\times [I_{i,n-1}\sigma_{{I_i},n-1} e^{j(\theta+\phi_i)}]^*
\end{align}
where $\sigma_{{V_i},n-1}$ is used to represent $\sigma_{V_i,n-1}(\{V_{i,n}V_{i,n}^*,I_{i,n}I_{i,n}^*\})$ for ease of notation, and similarly for $\sigma_{{I_i},n-1}$.
Considering $\mathcal{H}(\{V_{i}V_{i}^*,I_{i}I_{i}^*\})$ to be analytic in $D^C$ and applying Wirtinger's calculus \cite{amin2011wirtinger}, since 
\begin{align}
 \dfrac{\partial \mathcal{H}}{\partial V_{i,n-1}} = &\dfrac{\partial \mathcal{H}}{\partial V_{i,n-1}V_{i,n-1}^*}.\Bigg(\dfrac{\partial V_{i,n-1}V_{i,n-1}^*}{\partial V_{i,n-1}}\Bigg)\nonumber \\
 = & \dfrac{\partial \mathcal{H}}{\partial V_{i,n-1}V_{i,n-1}^*}.V_{i,n-1}^*,
\end{align} 
we have 
\begin{gather}
\sigma_{V_i,n-1}=\sqrt{\dfrac{1}{\eta_{i,n-1}}\Big(-\dfrac{\partial \mathcal{H}}{\partial  V_{i,n-1}}+\lambda V_{i,n-1}^*\Big)}
\end{gather}
where
\begin{gather}
\eta_i=V_{i,n-1}^*\sum\limits_{k=1}^N \Bigg(V_{k,n-1}\Big[-\dfrac{\partial \mathcal{H}}{\partial V_{k,n-1}}+\lambda V_{k,n-1}^*\Big]
\nonumber \\
+I_{k,n-1}\Big[-\dfrac{\partial \mathcal{H}}{\partial I_{k,n-1}}+\lambda I_{k,n-1}^*\Big]\Bigg)
\end{gather}
The discrete time update equations for $V_{i,n}$ is thus given by:
\begin{align}
{V_{i,n}} & \leftarrow V_{i,n-1}\sigma_{{V_i},n-1} e^{j\theta}
\end{align}
Similar expressions can also be derived for the current phasors.
\newline \hspace*{\fill} \qeda
\newline \textit{Lemma:} \textit{A continuous-time variant of the model given by Equation (\ref{eq_thm1}) is given by:}
\begin{align}
\label{eq_thm2}
\dfrac{\partial V_{i}(t)}{\partial t}=j\omega\sigma_{V_i}(t)V_{i}(t)-\Delta \sigma_{V_i}(t)V_i(t) .
\end{align}
\newline Proof: The difference equations for the voltage phasors are given by:
\begin{align}
{V_{i,n}}-V_{i,n-1}\leftarrow V_{i,n-1}\sigma_{{V_i},n-1} [e^{j\theta}-1]+V_{i,n-1}[\sigma_{{V_i},n-1} -1] 
\end{align}
which can be rewritten as:
\begin{align}
{V_{i,n}-V_{i,n-1}}\leftarrow V_{i,n-1}\sigma_{{V_i},n-1}{[\text{e}^{j\omega\Delta t}-1]}
-V_{i,n-1}\Delta\sigma_{V_i,n-1},
\end{align}
where $\Delta \sigma_{V_i,n-1}=1-\sigma_{V_i,n-1}$, and $\theta=\omega\Delta t$, where $\omega$ is the common oscillation frequency of all the phasors, and $\Delta t$ is the time increment. In the limiting case, when $\Delta t \rightarrow 0$, this reduces to the following continuous time differential equation for the complex variable $V_i(t)$:
\begin{align}
\dfrac{\partial V_{i}(t)}{\partial t}=j\omega\sigma_{V_i}(t)V_{i}(t)-\Delta \sigma_{V_i}(t)V_i(t),
\end{align}
\newline \hspace*{\fill} \qeda
\par In the steady state, since $H$ is Lipschitz continuous in $D$, $\sigma_{V_i}(t)\overset{t\rightarrow\infty}{\longrightarrow} 1$, the dynamical system above thus reduces to:
\begin{align}
\dfrac{\partial V_{i}(t)}{\partial t}=j\omega V_{i}(t) .
\end{align}
which implies that the steady state response of the complex variable $V_{i}(t)$ corresponds to steady oscillations with a constant angular frequency of $\omega$. 
\par The difference equations in terms of the nodal currents can be similarly written as:
\begin{align}
{I_{i,n}-I_{i,n-1}}\leftarrow I_{i,n-1}\sigma_{I_i,n-1}{[\text{e}^{j(\omega+\omega_{\phi_i})\Delta t}-1]} \nonumber \\
-I_{i,n-1}\Delta\sigma_{I_i,n-1},
\end{align}
where $\omega_{\phi_i}=\dfrac{d \phi_i}{d t}$. The equivalent continuous domain differential equation is then given by:
\begin{equation}
\dfrac{\partial I_i(t)}{\partial t} =j(\omega+\omega_{\phi_i})\sigma_{I_i}(t)I_{i}(t)-\Delta \sigma_{I_i}(t)I_i(t).
\end{equation}
\newline \hspace*{\fill} \qeda
\subsubsection{\textbf{Condition 3}}
 \textit{Considering $\beta \neq 0$ in Equation (\ref{eq_optimclass}), additional phase constraints can be incorporated in the dynamical system represented by using the update rules in Equations (\ref{eq_mainvol})-(\ref{eq_mainphase}).} 
\textit{In steady state, for $\lvert V_i \rvert ^2 \lvert I_i \rvert ^2 \neq 0$, the system settles to $\phi_i =\pm \pi/2 \enskip \forall i$. Additionally, for sufficiently small values of $\beta$ (or if $\beta$ is slowly increased during the optimization process), the system converges to the optimal point of the modified objective function $\mathcal{H}(\{\lvert V_i \rvert,\lvert I_i \rvert\})$.}
\newline Proof: Since $\mathcal{L}(\{\lvert V_i \rvert,\lvert I_i \rvert,\phi_i \})$ is Lipschitz continuous in both $\lvert V_i \rvert ^2$ and $\lvert I_i \rvert ^2$, the same form of the update equations proved in Lemma 2 can be applied. For arriving at the updates for the phase angles $\phi_i$, we will use a similar approach as shown in Equation (\ref{eq_twovar}). We can split $\phi_{i}$ as $\phi_{i}=\phi_{i}^{+}-\phi_{i}^{-}, \enskip \phi_{i}^{+},\phi_{i}^{-}>0$, which implies that $\phi_{i}^{+} +\phi_{i}^{-}=\pi$. 
We can then apply the growth transform dynamical system \cite{chatterjee2018decentralized} to get
\begin{align}
\text{and} \enskip \tau_i \omega_{\phi_i} +\phi_{i}(t) &=g_{\phi_i}(t),
\end{align}
where $\omega_{\phi_i}=\dfrac{d \phi_i(t)}{d t}$ and
\begin{align}
g_{\phi_{i}}= \pi \dfrac{\phi_{i}^{+}\Big(-\dfrac{\partial \mathcal{L}}{\partial \phi_{i}^{+}}+\lambda \Big)-\phi_{i}^{-}\Big(-\dfrac{\partial \mathcal{L}}{\partial \phi_{i}^{-}}+\lambda \Big)}{\phi_{i}^{+}\Big(-\dfrac{\partial \mathcal{L}}{\partial \phi_{i}^{+}}+\lambda\Big)+\phi_{i}^{-}\Big(-\dfrac{\partial \mathcal{L}}{\partial \phi_{i}^{-}}+\lambda\Big)}
\end{align} 
Since $\dfrac{\partial \mathcal{L}}{\partial \phi_{i}}=\dfrac{\partial \mathcal{L}}{\partial \phi_{i}^{+}}=-\dfrac{\partial \mathcal{L}}{\partial \phi_{i}^{-}}$, the above can be simplified to $g_{\phi_i}=\pi \dfrac{\lambda \phi_{i}-\pi \dfrac{\partial \mathcal{L}}{\partial \phi_{i}}}{-\phi_{i}\dfrac{\partial \mathcal{L}}{\partial \phi_{i}}+\lambda \pi}$ This implies that the voltage and current phasors corresponding to the $i-$th node in the network may be phase shifted by an additional amount $\phi_{i}$ with respect to the absolute reference. 
\par Since for optimality, $\phi_i = \pm \pi/2$ for $\lvert V_i \rvert ^2 \lvert I_i \rvert ^2 \neq 0$ in the steady state, the net energy dissipation in steady state is zero, i.e., $\sum \limits_{n=1}^N \lvert V_i \rvert ^2 \lvert I_i \rvert^2 \cos^2 \phi_i=0$. Also, in the steady state, for sufficiently small values of the hyperparameter $\beta$, 
\begin{align}
\underset{\{\lvert V_i \rvert,\lvert I_i \rvert,\phi_i\}} {\text{minimize}} \quad  \mathcal{L}(\{\lvert V_i \rvert,\lvert I_i \rvert,\phi_i\})= \underset{\{\lvert V_i \rvert,\lvert I_i \rvert\}} {\text{minimize}} \quad \mathcal{H}(\{\lvert V_i \rvert,\lvert I_i \rvert\}),
\end{align}
which implies that the system reaches the optimal solution with zero active power in the post-learning stage.
\newline \hspace*{\fill} \qeda


\end{document}